\pdfoutput=1
\documentclass[11pt]{article}

\usepackage{acl}

\usepackage{times}
\usepackage{latexsym}
\usepackage{graphicx}

\usepackage[T1]{fontenc}
\usepackage[utf8]{inputenc}

\usepackage{microtype}
\usepackage{multirow}
\usepackage{multicol}
\usepackage{soul}
\usepackage{multirow}
\usepackage{makecell}
\usepackage[utf8]{inputenc}
\usepackage{microtype}
\usepackage{graphicx}
\usepackage{soul}
\usepackage{float}
\usepackage{comment}
\usepackage{adjustbox}
\usepackage{booktabs}
\usepackage{amsmath}
\usepackage{cleveref}
\usepackage{mathabx, nicefrac}
\usepackage[shortlabels]{enumitem}
\usepackage{subcaption}
\usepackage{xcolor}

\usepackage{multicol}
\usepackage{caption}
\usepackage{subcaption}
\usepackage{color,soul}
\usepackage{microtype}
\usepackage{dblfloatfix}

\usepackage[ruled,linesnumbered]{algorithm2e}
\SetKwInput{KwInput}{Input}
\SetKwInput{KwOutput}{Output}

\newcommand{\numglue}{\textsc{NumGLUE}}

\newcommand{\etc}{\textit{etc.}}
\newcommand{\eg}{e.g.}
\newcommand{\ie}{i.e.}
\newcommand{\cha}[1]{{#1}}
\title{Towards Question Format Independent Numerical Reasoning: A Set of Prerequisite Tasks}
\title{NumGLUE: A Suite of Fundamental yet Challenging Task for GPT3}
\title{\numglue: A Suite of Fundamental yet Challenging \\ Mathematical Reasoning Tasks}

\author{
Swaroop Mishra$^{1}$ $\;$ Arindam Mitra$^{2}$ $\;$ \textbf{Neeraj Varshney}$^{1}$ 
$\;$ \textbf{Bhavdeep Sachdeva}$^{1}$ $\;$ \\
\textbf{Peter Clark}$^{3}$ $\;$ 
\textbf{Chitta Baral}$^{1}$ $\;$ \textbf{Ashwin Kalyan}$^{3}$ $\;$ 
\\\\
 $^1$Arizona State University \; $^2$Microsoft Research \; 
 $^3$Allen Institute for AI
}
\begin{document}
\maketitle
\begin{abstract}
    % Given the ubiquitous nature of numbers in text, reasoning with numbers to perform simple calculations is an important skill of AI systems. 
    % While many datasets and models have been developed to this end, state-of-the-art AI systems are brittle; failing to perform the underlying mathematical reasoning when they appear in a slightly different scenario.
    % Drawing inspiration from GLUE that was proposed in the context of natural language understanding, we propose NumGLUE, a multi-task benchmark that evaluates the performance of AI systems on eight different tasks, that at their core require simple arithmetic understanding.
    % We show that this benchmark is far from being solved with neural models including state-of-the-art large-scale language models performing significantly worse than humans (lower by 46.4 %). 
    % Further, NumGLUE promotes sharing knowledge across tasks, especially those with limited training data as evidenced by the superior performance (average gain of 3.4 % on each task) when a model is jointly trained on all the tasks as opposed to task-specific modeling.
    % Finally, we hope that NumGLUE will encourage systems that perform robust and general arithmetic reasoning within language, a first step towards being able to perform more complex mathematical reasoning.
    Given the ubiquitous nature of numbers in text, reasoning with numbers to perform simple calculations is an important skill of AI systems. 
    While many datasets and models have been developed to this end, state-of-the-art AI systems are brittle; failing to perform the underlying mathematical reasoning when they appear in a slightly different scenario.
% These systems while ``solving'' the specific dataset fail to truly understand the core reasoning skill of simple arithmetic.
    Drawing inspiration from GLUE \cite{wang2018glue} that was proposed in the context of natural language understanding, we propose \numglue, a multi-task benchmark that evaluates the performance of AI systems on eight different tasks, that at their core require simple arithmetic understanding.
    We show that this benchmark is far from being solved with neural models including state-of-the-art large-scale language models performing significantly worse than humans (lower by 46.4\%). 
    Further, \numglue~promotes sharing knowledge across tasks, especially those with limited training data as evidenced by the superior performance (average gain of 3.4\% on each task) when a model is jointly trained on all the tasks as opposed to task-specific modeling.
    Finally, we hope that \numglue~ will encourage systems that perform robust and general arithmetic reasoning within language, a first step towards being able to perform more complex mathematical reasoning\footnote{
% https://instructions.apps.allenai.org/
\url{https://allenai.org/data/numglue}
}.
\end{abstract}

\section{Introduction}

Reasoning with numbers is an important skill that occurs in various day-to-day scenarios and not surprisingly, numbers are ubiquitous in textual data. 
To train AI reasoning systems that can perform simple mathematical reasoning, many tasks have been proposed \citep{dua2019drop, ravichander2019equate, koncel2016mawps}.
Despite these efforts, current state-of-the-art AI systems are brittle and fail when problems involving similar mathematical reasoning is posed in a slightly different manner. 
For instance, presenting a word problem in a different manner as shown in \cref{fig:cover}, while hardly affecting human performance, is sufficient to confuse state-of-the-art AI systems\footnote{The recently released GPT3-Instruct, a fine-tuned model with 175B parameters produces inconsistent answers for these questions. See supplementary material: GPT3-Instruct's Response for more details.}.
This brittleness in reasoning indicates that the models latch on to spurious signals in the specific dataset resulting in ``solving'' the dataset while not truly understanding the underlying reasoning skill of simple arithmetic.
\begin{figure}
\fbox{
    \parbox{0.96\columnwidth}{
\underline{Original Word Problem} \\ 
\emph{John had 5 apples. He gave 3 to Peter. How many apples does John have now?} \\ \\
\underline{Fill In The Blanks Format} \\
    John had 5 apples. He gave 3 to Peter. John has \rule{1cm}{0.15mm} apples now. \\ \\
\underline{NLI Format} \\ 
    Premise: John had 5 apples. He gave 3 apples to Peter. Hypothesis: John has 2 apples now. Does the hypothesis entail, contradict or is neutral to the premise?  \\ \\
\underline{Comparison Format} \\
    John had 5 apples. He gave 3 to Peter. Who has more apples?
}}
\vspace{-5pt}
\caption{A system that can robustly perform numeric reasoning over language should be able to solve problems such as the above, regardless of how the problem is posed. However, we observe existing systems are brittle; producing inconsistent solutions to such minor stylistic variations.}\label{fig:cover}
\vspace{-15pt}
\end{figure}
Further, we believe that building AI systems that can truly understand and apply simple arithmetic reasoning is a mandatory first step towards successfully tackling complex mathematical reasoning skills \cite{saxton2019analysing, hendrycks2020measuring, hendrycks2021measuring}.
%%%%%%%%%%%%%%%%%%%%%%%%%%%
%%%%% FIGURE
%%%%%%%%%%%%%%%%%%%%%%%%%%%
% \begin{figure*}[t]
% \centering
  % \includegraphics[scale=0.39, trim=3cm 0cm 0cm 0cm]{sections/test2png (2).png}
% \caption{Performance of zeroshot, fewshot and finetuning baselines (Section \ref{exp}) across NumGLUE. There is a signficant gap between the highest performing model and the human baseline.}
% \label{fig:teaser}
% \end{figure*}
% \begin{figure*}[t]
% \centering
% \begin{minipage}{\textwidth}
  % \centering
  % \includegraphics[width=0.78\textwidth]{sections/NumGLUE2.pdf}
% %   \captionof{figure}{A figure}
% %   \label{fig:test1}
% \caption{Our dataset NumrGLUE (center in the yellow circle) has been positioned with respect to existing datasets. T1-T8 represents 8 types of questions. Note that, NumGLUE contains the feature of being format invariant unlike other datasets. Position of datasets within clusters is done based on their semantic category, for example T1 Numerical Commmonsense QA is closer to the cluster of Commonsense Reasoning + Knowledge of Facts; its position reflects the same }
% \label{fig:map}
% \end{minipage}%
% \end{figure*}
%%%%%%%%%%%%%%%%%%%%%%%%%%%
%%%%% FIGURE
%%%%%%%%%%%%%%%%%%%%%%%%%%%
\\ \\
\textbf{NumGLUE.} To this end, we propose \numglue, a multi-task benchmark consisting of eight different tasks that at their core test for arithmetic reasoning skills.
For example, as discussed in \cref{fig:cover}, tasks can involve word problems presented in a slightly different manner or can involve additional reasoning strategies like commonsense reasoning or reading comprehension to be combined with the core skill of simple arithmetic.
Our benchmark consists of four new tasks in addition to four existing ones; with ${\sim}100K$ problems spread across eight differet tasks.
The motivation behind \numglue~ is similar to GLUE \cite{wang2018glue, wang2019superglue}, a multi-task benchmark that aimed at models that demonstrated superior language understanding by learning the underlying linguistic features. 
\numglue~is designed with goal of progressing towards AI systems that are capable of performing arithmetic reasoning in a general setting; achieving superior performance on our benchmark requires the ability to correctly identify and perform the underlying arithmetic reasoning without relying on task or dataset-specific signals.
Finally, we hope that \numglue~ will encourage systems that perform robust and general numeric reasoning within language, a first step towards being able to perform more complex mathematical reasoning.
% \\ \\
% \textbf{AI for Basic Arithmetic.} Despite the impressive achievements of AI systems in various fields including challenging settings like game-playing (\ak{cite AlphaGO, DeepBlue, etc}), we are far from reliably performing the fundamental task of arithmetic. 
% \ak{For instance, state-of-the-art large scale language models like GPT-3 attain only xx.yy\% on the task of ABC that primarily involves understaning simple English sentences to output a simple arithmetic formula.}
% While the mathematical reasoning community in general is working on challenging tasks that require solving hard mathematical problems like combinatorics, differential equations or number theory ak{cite them}, we believe that the inability of models to understand the more foundational skill of arithmetic is a crucial bottleneck in our quest for building AI that can perform complex mathematics.
% \\ \\
% \textbf{Going Beyond Individual Datasets.} As mentioned previously, basic arithmetic is a crucial skill that occurs in multiple forms and scenarious -- \eg simple word problems, questions that require commonsense knowledge (\eg how many hands does a human have?) in addition to perform simple arithmetic.
% Therefore, in order to build AI systems that truly understand basic arithmetic, we need a system that performs well across multiple datasets and language formats.
% In addition, such an evaluation setting, allows us to understand if an AI system is not merely overfitting to spurious biases in a specific dataset to achieve superior performance.
\\ \\
\textbf{Contributions.} 
\begin{enumerate}[topsep=4pt, leftmargin=12pt, itemsep=0pt]
    \item We introduce \numglue -- a multi-task benchmark consisting of eight different tasks, including 4 new ones, whose solution at its core requires an understanding of simple arithmetic. 
        % The meta-dataset is intended to be solved by a single model that captures the underlying reasoning skill of arithmetic understanding.
    \item We demonstrate that \numglue~ is a challenging benchmark even for state-of-the-art large scale language models, obtaining poor scores not only in zero or few shot settings but also after fine-tuning.
        This indicates a fundamental barrier for AI systems; one that needs to be breached before complex mathematical challenges can be successfully tackled. 
    \item Finally, we propose a memory-augmented neural model to demonstrate the utility of such a multi-task meta dataset. 
        Our proposed model when trained on the entirety of \numglue~ obtains an average improvement of 3.4\% on each task as opposed to task-specific training -- indicating that joint training leads to beneficial transfer owing to the common theme of arithmetic reasoning.
\end{enumerate}

% Numerical Reasoning datasets ordered in increasing order of complexity, our proposed dataset \sm{NumerGLUE} is at the bottom as it is least complex. Complexity here refers to the mathematical knowledge and number of mathematical operations required to solve a task. E.g. Datasets at the top require probability, statistics and other mathematical knowledge beyond commonsense, so they are in the top of the list
% \cite{hendrycks2021measuring}
% \cite{saxton2019analysing}
% \cite{polu2020generative}
% \cite{ling2017program}
% \cite{zhang2020machine}
% \cite{seo2015solving}
% \cite{mitra2015learning}
% \cite{amini2019mathqa}
% \cite{zhang2020language}

\section{Related Work}

\textbf{Datasets for Numerical reasoning.}
Quantitative reasoning has been a challenging problem for a long time. Small question answering datasets were proposed to understand the quantitative aspect of natural language such as the template-based dataset which solved questions with equations as parameters \cite{kushman2014learning}, addition-subtraction dataset \cite{Hosseini14learningto} and arithmetic problems dataset \cite{koncel2015parsing}. Difficulty of questions were increased in subsequent datasets \cite{roy2016solving}, \cite{upadhyay2016learning}. Later, larger datasets were created to facilitate deep learning research \cite{ling2017program,dua2019drop}. Several other maths datasets have been proposed to improve explainability \cite{amini2019mathqa}, diversity \cite{miao-etal-2020-diverse}, scale information in language embeddings \cite{zhanglanguage} and hardness of math questions \cite{hendrycks2021measuring}.
\\ \\
One of the motivations behind creating this benchmark is to test for simple arithmetic reasoning independent of the context or the presentation style of the problem. 
Further, To the best of our knowledge, our work is the first to consider multiple tasks in the numerical reasoning space. 
\\ \\
\textbf{Multi-Task Benchmarks.} With increased success of deep learning based models on individual tasks, there has been a significant push both in the NLP community and in the broader AI community towards general purpose models that excel at multiple tasks. 
Naturally, various benchmarks and challenges that test for such understanding have been proposed. 
For instance, the BAbI dataset \cite{weston2015towards}, GLUE \cite{wang2019superglue} and the subsequent harder SuperGLUE \cite{wang2019superglue} were proposed to both evaluate and drive progress in language understanding via shared linguistic knowledge across tasks.
% McCann~\etal\cite{McCann2018decaNLP}
\citet{McCann2018decaNLP} build a multi-task dataset via a novel approach -- formatting each task as that of question-answering. 
In the more restricted setting of reading comprehension, \citet{dua2019orb} and \citet{downeygetting} build a meta-dataset that spans multiple domains and reasoning skills.
\\ \\
\textbf{Multi-task Models.} With the growing interest towards models that go beyond specific datasets, various neural models that can perform mutliple tasks have been proposed. 
When the underlying reasoning is similar -- eg. commonsense reasoning, problem decomposition or linguistic understanding -- it has been found that training on multi-task datasets yields more robust and accurate models.
For instance, the Multi-task Question Answering Network \cite{McCann2018decaNLP}, T5 \cite{raffel2019exploring}, GPT3 \cite{NEURIPS2020_1457c0d6} and GPT3-Instruct models aim to build general purpose language models that are capable of transferring linguistic understanding across tasks.
A similar approach is taken by \citet{khashabi2020unifiedqa} in the setting of question-answering and \citet{lourie2021unicorn} in the scope of commonsense reasoning.
Further, Muppet \cite{aghajanyan2021muppet} adds an additional step of pre-finetuning between pretraining and finetuning that improves generalization to multiple tasks. 

\begin{table*}[t]
    \small
    \centering
    % \resizebox{\columnwidth}{!}{%
    \begin{tabular}{p{1cm}p{4.64cm}p{0.5cm}p{8.3cm}}
    \toprule
        Task & Question Setting & Size & Example \\ 
        \\
        \midrule
         \multirow{2}*{\textsc{Task 1}} &  \multirow{2}*{Commonsense + Arithmetic} & \multirow{2}*{404}& Question: A man can lift one box in each of his hands. How many boxes can a group of 5 people hold in total? Answer: 10 \\
 \midrule
         \multirow{2}*{\textsc{Task 2}} &  \multirow{2}*{{Domain specific + Arithmetic}} &  \multirow{2}*{1620} & Question: How many units of $H_2$ are required to react with 2 units of  $C_2H_4$ to form 2 units of  $C_2H_6$? Answer: 2 \\

\midrule
        \multirow{5}*{\textsc{Task 3}} &  \multirow{5}*{{Commonsense + Quantitative}} &  \multirow{5}*{807} & Question: A person wants to get shopping done quickly. They know that they can get through the check-out at big store in 5 minutes whereas it can take 20 minutes at small store. The store they go to finish quickly is? (A) big store (B) small store? Answer: big store\\
\midrule
\multirow{3}*{\textsc{Task 4}} & \multirow{3}*{{Fill-in-the-blanks}} &  \multirow{3}*{1100}& Question: Joan found 70 seashells on the beach. She gave Sam some of her seashells. She has 27 seasshells left. She gave \_\_\_\_\_ seashells to Sam? Answer: 43\\
\midrule
\multirow{2}*{\textsc{Task 5}} &  \multirow{2}*{{RC + Explicit Numerical Reasoning}} & \multirow{2}*{54212} & Passage: <>. Question: How many counties were added in 1887? Answer: 2\\
\midrule
\multirow{2}*{\textsc{Task 6}} &  \multirow{2}*{{RC + Implicit Numerical Reasoning}} &  \multirow{2}*{32724}& Passage: <>. Question: Which player kicked the shortest field goal? Answer: David Akers\\
\midrule
\multirow{3}*{\textsc{Task 7}} &  \multirow{3}*{Quantitative NLI} &  \multirow{3}*{9702}& Statement 1: James took a 3 - hour bike ride,
  Statement 2: James took a more than 1 - hour bike ride,
  Options: Entailment or contradiction or neutral?,
  Answer: Entailment\\
\midrule
\multirow{2}*{\textsc{Task 8}} &  \multirow{2}*{Arithmetic word problems} & \multirow{2}*{1266} & Question: Joe had 50 toy cars. If he gives away 12 cars, how many cars will he have remaining?,
  Answer: 38\\
\bottomrule

    \end{tabular}
    % }
    \caption{Size and example of each task in the NumGLUE benchmark. RC: Reading Comprehension}
    \label{tab:size_table}
    % \vspace{-3mm}
\end{table*}

\section{\numglue}\label{createdata}

% \begin{figure}
% \centering
% \begin{minipage}{.5\textwidth}
%   \centering
%   \includegraphics[width=\linewidth,height=8cm,keepaspectratio]{LaTeX/NumericalReasoningType2Flow.png}
% %   \captionof{figure}{A figure}
% %   \label{fig:test1}
% \caption{Flow 2}
% \label{Flow2}
% \end{minipage}%
% \end{figure}

% \begin{figure}
% \centering
% \begin{minipage}{.5\textwidth}
%   \centering
%   \includegraphics[width=\linewidth,height=8cm,keepaspectratio]{LaTeX/NumericalReasoningType4Flow.png}
% %   \captionof{figure}{A figure}
% %   \label{fig:test1}
% \caption{Flow 4}
% \label{Flow4}
% \end{minipage}%
% \end{figure}
% \begin{figure*}
%     \centering
%     \includegraphics[width=16cm]{Pictures/Intro diagram.png}
%     \caption{In the conventional approach}
%     \label{fig:Intro_diagram}
% \end{figure*}

% Our dataset consists of a variety of question types that involve numerical computation and reasoning. 
%  We carefully select eight setting that involve numerical reasoning such that each setting has distinct properties and one is not just an extension of others. 
As mentioned previously, our \numglue~benchmark consists of both new and already existing arithmetic reasoning tasks. 
We first begin by introducing the novel datasets curated by us before providing a brief overview of existing tasks that are part of \numglue. 
Finally, in this section, we provide an analysis of the datasets demonstrating that it contains interesting and diverse linguistic and mathematical properties. 
% We divide our dataset into two broad categories. First category includes novel datasets and second is a collection of existing datasets.
\\ \\
\noindent \textbf{\numglue~ Benchmark.}
Our proposed \numglue~benchmark is a collection of eight different tasks that together include ${\sim}100K$ questions. 
The tasks may either be self-contained or require additional background knowledge (\eg commonsense reasoning) to arrive at the final solution; however, all the tasks, at their core, involve arithmetic reasoning.
\Cref{tab:size_table} shows an example question belonging to each task along with indicating the total number of data points associated with each task. 
It is important to note that tasks are imbalanced with only ${\sim}400$ examples for Task 1 and nearly $50K$ questions under Task 5. 
While we could have under-sampled the questions to create a balanced suite, we retain the imbalanced dataset in order to mimic the real world -- for instance, arithmetic word problems are more abundant as opposed to word problems that may require commonsense reasoning in addition to arithmetic reasoning.
\\ \\
% In a real world setting, number of problems in each type of data is different. Instead of under-sampling or over-sampling data across types, we decide to keep them disproportionately to mimic the real world setting. 
\textbf{Data Partition and Evaluation.} We randomly partition data in each task into training (70\%), development (10\%) and test (20\%) sets . 
In the case of reading comprehension tasks (Task 5 and 6), we assign all questions corresponding to a passage to the same split -- we do this in order to discourage any data leakage and thereby, allowing models to potentially rely on memorization to arrive at the correct answer. 
\\ \\
For each task, we report the F1 measure and as an aggregate measure of performance on the \numglue~benchmark similar to ~\citet{dua2019drop}, we report the (unweighted) average of the F1 scores corresponding to each task. 
\\
% \sm{Can you about evaluation metrics -- F1 and \numglue~score?}
% We ensure that there is no data leakage among these splits. In order to ensure this for RC problems, we keep all questions of a passage in only one of the splits. Similarly, in other setting, questions which are very similar to each other are kept in only one of the splits. 
% need to change 
% This way, we reduce the possibility of memorization by language models.
% Table \ref{tab:size_table} enlists size and example of each question type in the NUMBERGAME benchmark.
%add spurious paper as the reference here.

\subsection{Novel Datasets}

The novel tasks proposed as part of \numglue~are a combination of both freshly collected data and intelligent modifications of already existing datasets. The four novel arithmetic reasoning tasks introduced are as follows \footnote{We annotate the datasets manually. We provide the exact flow used to generate questions of each task in the supplementary materials: Construction of \numglue.}: \\ \\
% This consists of datasets which we have created manually by hand. Our motivation behind manual data creation was to create high quality data, as automated methods and crowdsourcing approaches have produced dataset artifacts that cause overfitting and prevent generalization \cite{gururangan2018annotation, bras2020adversarial}.
% Recently, a few QA datasets have been proposed that require external knowledge to answer the questions. 
% Out of four datasets which we have created in this category, three of them require knowledge, specifically numerical common sense knowledge. 
% We create data in four different setting. Three of them require knowledge, specifically numerical common sense knowledge and the fourth one is a collection of questions in completion format. None of the knowledge is explicitly provided in the question. Table \ref{tab:sample1} shows some samples of this category. 
% We also create adversarial samples of each type to improve robustness of our dataset.
% Based on the kind of knowledge required by the questions, we divide this category further into four sub-categories.
\noindent \textbf{Task 1: Commonsense + Arithmetic Reasoning.}
Consider the following question -- \emph{How many faces do 10 dice have?} Answering this not only requires simple arithmetic \ie multiplying the number of faces in a die by ten but also requires knowing that a standard die has six faces. 
We collect this dataset by first asking the annotator to write down a numerical commonsense fact (\eg a human has 2 hands, a day has 24 hours \etc) and then use frame a question that requires using this numerical fact as part of a simple arithmetic calculation. 
% We present the exact flow used to generate such questions in the supplementary materials: Construction of NumGLUE.
\\ \\
\noindent \textbf{Task 2: Domain Specific + Arithmetic Reasoning.}
\emph{How many units of hydrogen are required to produce 10 units of water?} 
This question, similar to the previously introduced task of arithmetic reasoning questions, requires additional domain-specific knowledge -- specifically, that each unit of water contains two units of hydrogen. 
We curate a dataset of such questions that require both domain-specific knowledge and arithmetic reasoning motivated by the finding that QA systems perform poorly on the ARC dataset \citet{clark2018think} consisting of grade-school level science questions.
Specifically, the dataset collected by us requires understanding of a small set of chemistry (conservation of mass in chemical reactions) and physics principles ($speed = \nicefrac{distance}{time}$).
\\ \\
\noindent \textbf{Task 3: Commonsense + Quantitative Comparison.}
\emph{A golf ball weighs 40g and a baseball weighs 150g. Which has a higher gravitational force?}
Answering this question requires both knowing that mass is directly proportional to gravitational force and a numerical comparison via subtraction. 
We collect such quantitative comparison questions by using the QuaRel dataset \cite{tafjord2019quarel} containing questions from diverse fields such as physics and economics as the starting point.
The annotator chooses a subset of these questions that involve numerically comparable quantities (for instance, in this example, mass of the objects involved) to create the required task of quantitative comparison questions. 
\noindent \textbf{Task 4: Fill-in-the-blanks Format.}
% MAWPS is collection of Math word problems and incorporates data from the earlier datasets like AddSub(), SingleOp(), MultiArith().
Unlike the previously proposed tasks that require external information (\eg commonsense knowledge) in addition to simple arithmetic reasoning, this task is self-contained but a stylistic variant of existing math word problems. 
We source word problems from the Arithmetic Word Problem repository \cite{roy2016solving, roy2017unit, roy2018mapping} and convert them into the fill-in-the-blanks format. 
For an example of such a conversion, refer to \cref{fig:cover}. 

\subsection{Existing Datasets}

We now review existing datasets while discussing any modifications made when including them in \numglue. 
In general, for all the datasets included, we perform a filtering step to clean and control for the quality of the data points being included. 
This step includes -- a) discarding questions that do not have answer annotations b) eliminating questions with high lexical overlap with the remainder of the dataset and c) fixing any type mismatches present in the data (\eg ``7.0 students'' $\rightarrow$ ``7 students''). 
% We compile questions from some of the existing datasets that involve numerical reasoning. In order to avoid adding repetitive questions and ensure high quality of our dataset, we filter questions programmatically following a four step procedure. First, we remove questions that do not have annotated answers. Second, we remove grammatically incorrect questions. Then, we eliminate problems which have high lexical overlap (implemented using Semantic Textual Similarity) with rest of the dataset, thus ensuring that our dataset incorporates mostly unique concepts. Next, we rectify type mismatch issues such as, "there are 7.0 students" to "there are 7 students" as the number of students is not a float quantity. We also manually investigate several random samples and discard data where the annotation is incorrect. We compile a total of four question types from the existing datasets.
%DROP \cite{dua2019drop}, EQUATE \cite{ravichander2019equate}
\\ \\
\noindent \textbf{Task 5: Reading Comprehension (RC) + Explicit Numerical Reasoning.}
We select a subset from the DROP \cite{dua2019drop} dataset to create this task.
Specifically, the selected questions involve reading comprehension and numerical reasoning but importantly, the required answer is also a number. 
\\ \\
\textbf{Task 6: Reading Comprehension (RC) + Implicit Numerical Reasoning.} 
Consider the following question based on a relevant passage -- \emph{Which state has the highest income tax rate?} Here, while the final answer is a name, arriving at it requires performing comparison (\ie subtraction). 
We classify such questions in the DROP dataset as a separate task in \numglue.
\\ \\
% Here the answer is not a numerical value but some sort of mathematical operation such as counting or sorting is required to answer the question. This category is inspired from the fact that many a times in real world some sort of math is implicitly required to answer a question.
%Here, we have the reading comprehension questions which require implicit math for answering. By implict math, we mean that answer here is not a number, but we have to do some sort of mathematical operation like comparison of two numbers to find the answer. 
%This category is inspired from the fact that many a times in real world we need some sort of math implicitly and accurately to answer a question.
\noindent \textbf {Task 7: Quantitative NLI}
EQUATE \cite{ravichander2019equate} introduces quantitative NLI questions that require simple arithmetic calculations to be performed in order to accurately classify the relationship between the provided premise and the hypothesis.
As noted in \cref{fig:cover}, many word problems can also be easily converted to this format and is therefore, a diverse and interesting task for evaluating arithmetic reasoning skills of AI systems. 
% Natural Language Inference (NLI) or Recognizing Textual Entailment (RTE) has been considered as a benchmark task in natural language understanding. Recently introduced EQUATE \cite{ravichander2019equate} dataset has quantitative NLI problems combined from diverse sources. We use EQUATE to add NLI questions to our dataset.%and process them using our processing techniques.
% add quartz's numbere problems, same for quarel
\\ \\
\textbf{Task 8: Arithmetic Word Problems}
Finally, we arrive at one of the earliest and extensively studied class of arithmetic reasoning problems \ie word problems. 
The specific dataset included as part of our \numglue benchmark is a combination of multiple datasets proposed by \citet{koncel2016mawps}, \cite{koncel2015parsing} and \citet{kushman2014learning}.
Further, to ensure that the benchmark as a whole is diverse, we eliminate questions that have a high sentence similarity with questions from the fill-in-the-blanks task.
\begin{figure*}[t]
\centering
   \includegraphics[scale=0.32, trim=3cm 0cm 0cm 0cm]{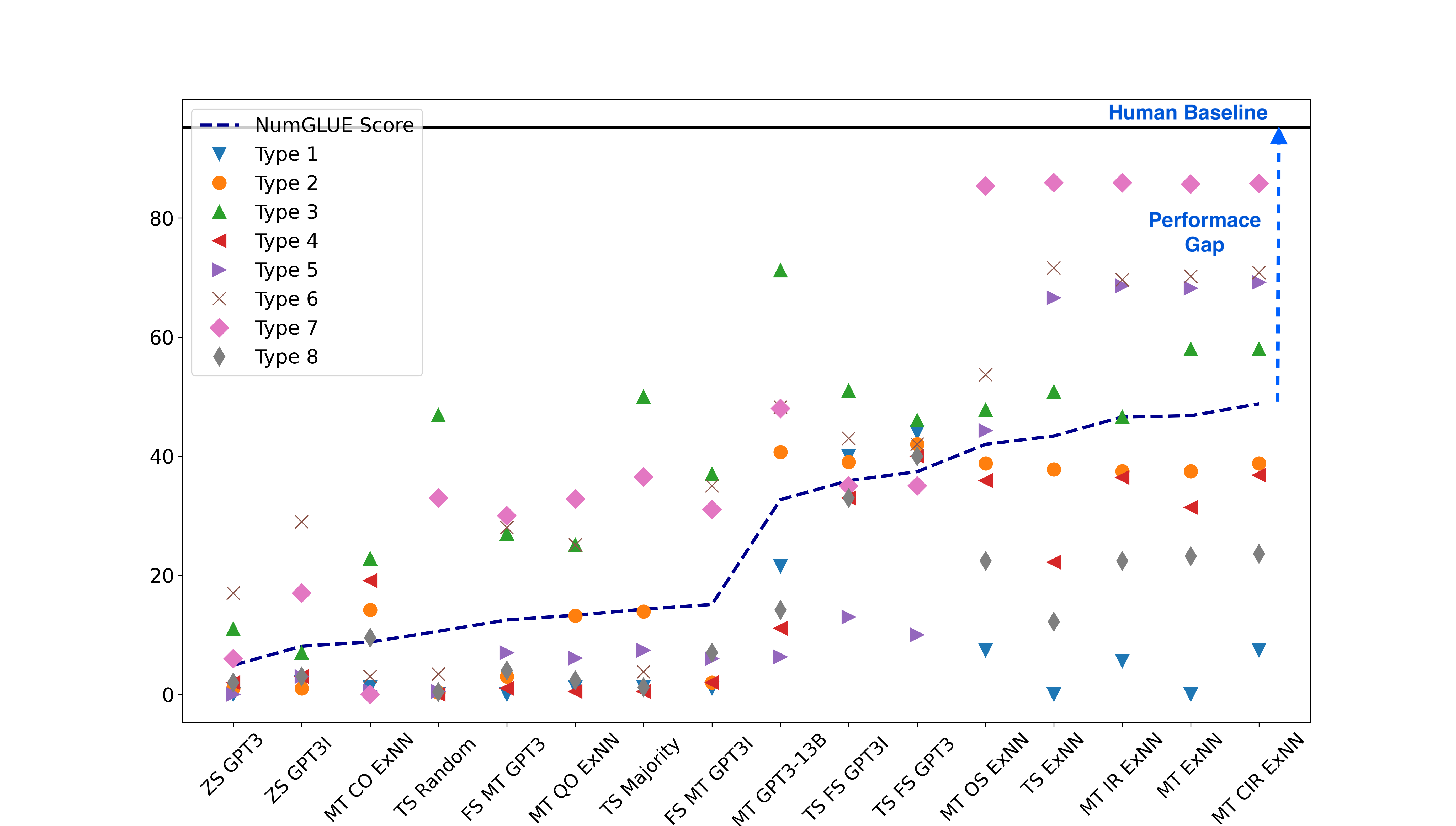}
\caption{Performance of zeroshot, fewshot and finetuning baselines (Section \ref{exp}) across NumGLUE. There is a signficant gap between the highest performing model and the human baseline. ZS: Zeroshot, GPT3I: GPT3-Instruct, MT: Multi-task, TS: Task-specific, QO: Question Only, CO: Context Only, EXNN: Ex-NumNet,FS: Few-shot, OS: Oversampling, IR: Information Retrieval, CIR: Conditional Information Retrieval.}
\label{fig:teaser}
\end{figure*}
\subsection{Data Quality Analysis:}
In order to ensure a high-quality test set, three independent annotators evaluate each question in the test set across all tasks. 
A tiny porton of the data marked as invalid or with disagreement between the annotators was excluded, resulting in a verified, high-quality \numglue~evaluation suite. We also  perform a variety of analysis and find that the novel question tasks we created (task 1-4) have higher quality than the existing question tasks since they have higher average vocabulary (number of unique words per number of samples), higher number of unique nouns, verbs and other POS tags and have less semantic textual similarity among each other (indicating lower repetition). Detailed analysis can be found in the supplementary material: Data Quality Analysis of \numglue.

\begin{figure*}[t]
\centering
\begin{minipage}{\textwidth}
  \centering
  \includegraphics[width=1\textwidth]{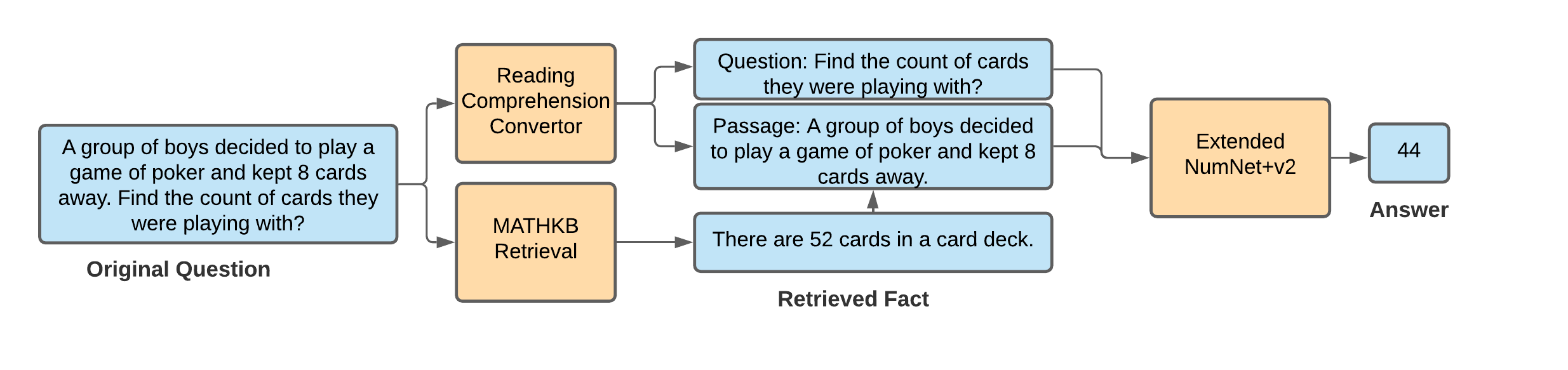}
\caption{Our proposed memory-augmented model that detects the type of task (1-8), uses Information Retrieval from \textit{MATH KB} and append the information that gets fed to Ex-NumNet}
\label{fig:knowledgemodel}
\end{minipage}%
\end{figure*}
\begin{table*}[]
    \small 
    \centering
    \resizebox{\linewidth}{!}{
\begin{tabular}{cccccccccccc}
\toprule 
Learning & Baseline & Baseline & Task 1 & Task 2 & Task 3 & Task 4 & Task 5 & Task 6 & Task 7 & Task 8 & NumGLUE\\ 
 & category & name&  & & & & & & & & Score\\
\midrule
\multirow{2}{*}{\textsc{Heuristic}} & Task-specific & Random & 0 & 0.3 & 46.9 & 0 & 0.5 & 3.4 & 33 & 0.4 & 10.6\\
& Task-specific & Majority & 1.2 & 13.9 & 50 & 0.5 & 7.4 & 3.8 & 36.5 & 1.2 & 14.3\\
\midrule
\multirow{2}{*}{\textsc{Zero-Shot}} & - & GPT3 & 0      & 1      & 11    & 2   & 0    & 17    & 6   & 2 & 4.9 \\
& - & GPT3-Instruct & 2      & 1      & 7    & 3   & 3    & 29    & 17    & 3 & 8.1   \\
\midrule
\multirow{4}{*}{\textsc{Few-Shot}} & Task-specific & GPT3 & \textbf{44}       & \textbf{42}        & 46      & 40     & 10  &42     & 35       & 40  & 37.4 \\
& Task-specific & GPT3-Instruct & 40       & 39        & 51      & 33     & 13  &43     & 35       & 33 & 35.9\\
% \midrule
& Multi-task & GPT3 & 0       & 3        & 27      & 1     & 7  &28     &30       & 4 & 12.5 \\
& Multi-task & GPT3-Instruct & 1       & 2        & 37      & 2     & 6  &35     & 31       & 7 & 15.1\\
\midrule
\textsc{Fine-tuning} & Multi-task &  GPT3-13B & 21.5 & 40.7 & \textbf{71.2}       & 11.1          & 6.3           & 48.2          & 48.0          & 14.2 &32.7\\
%phy:10%
\midrule
\multirow{7}{*}{\textsc{Fine-tuning}} & Multi-task (Q-only) & Ex-NumNet & 1.2          & 13.2          & 25.1       & 0.5            & 6.1              & 25.1          & 32.8          & 2.4 & 13.3\\
 & Multi-task (C-only) & Ex-NumNet & 1.2          & 14.2          & 22.8          & 19.1          & 0.6            & 3              & 0                & 9.5 & 8.8\\
 & Single-task &  Ex-NumNet& 0 & 37.8 & 50.8  & 22.2          & 66.6            & \textbf{71.6}          & 85.9          & 12.2 & 43.4 \\
 & Multi-task &  Ex-NumNet& 0              & 37.5          & 58  & 31.4          & 68.2            & 70.2 & 85.7          & 23.2& 46.8\\
% \midrule
& Multi-task + IR &  Ex-NumNet& 5.6          & 37.5          & 46.6       & 36.4 & 68.6 & 69.6          & \textbf{85.9} & 22.4 & 46.6 \\
 & Multi-task + CIR &  Ex-NumNet&           7.4  &         38.8 &      58 & \textbf{36.8} &  \textbf{69.2} &      70.8     & 85.8 & \textbf{23.6} & \textbf{48.8}\\
& Multi-task + OS &  Ex-NumNet& 7.4 & 38.8 & 47.8       & 35.9          & 44.3            & 53.7          & 85.4          & 22.4 &42.0\\
\midrule

- & - & Human & 94.4        & 94.5          & 97.8       & 95              & 94.7          & 96.1            & 96.5          & 92.8 &95.2\\
\bottomrule
\end{tabular}
}
 \caption{
        F1 performance of various baselines on the NumGLUE test set across various tasks 1-8. Human performance was calculated on 100 samples of each task (81 of Task 1) {[}*IR = Information Retrieval, CIR=Conditional Information Retrieval, OS=Oversampling, Q. Only: Question Only, C. Only: Context Only {]}.
    }
    \label{tab:main}
\end{table*}

\section{Experiments}
\label{exp}
In this section, we establish multiple baselines on our benchmark and discuss their performance. 
\subsection{Baselines}
We evaluate several baselines on our benchmark -- (i) Heuristic, (ii) Zero-shot, (iii) Few-shot, (iv) Fine-tuning and (v) Human. 
We use two kinds of model architectures (i) Neuro-symbolic, a memory augmented \emph{novel} architecture that extends Numnet+v2 \cite{ran2019numnet} and (ii) End-to-end, GPT3 \cite{NEURIPS2020_1457c0d6}. 
% For Neuro-symoblic, we extend NumNet+v2 \cite{ran2019numnet}, the best one among the existing models for which code is publicly available and develop Extended NumNet+v2. We use GPT3 \cite{NEURIPS2020_1457c0d6} as our end-to-end model.
\\ \\ 
\textbf{Architectures.} In the multi-task setting where the same model is trained on all the \numglue~tasks, we use Reading Comprehension (RC) as the common format -- converting each task to RC format via a set of hand-coded rules \footnote{More details in the supplementary material: Ex-NumNet}. 
In addition to being capable of faithfully representing all the constituent tasks, the RC format also allows us to inject additional context in the IR setting without affecting the rest of the pipeline \footnote{Henceforth we will be calling our extension to Numnet+v2 as Ex-NumNet}.
On the other hand, GPT3 being a generative model does not require such modifications. Importantly, note that both models are inputted the exact same information for the multi-task experiments.
% We use a numeracy-focused (macro-averaged) F1 score as evaluation metrics, similar to prior works in RC setting \cite{dua2019drop}.
\\ \\
\noindent \textbf{Heuristic Baselines with Task Oracle.}
For this baseline, we assume a task oracle that knows the task a particular question belongs (in a multi-task setting) -- we use this to make our heuristic baselines more competitive. 
% We assume that there is a type oracle that knows the question type. We add this to our heuristic baseline, since use of a single baseline across all eight types is not appropriate. 
The first heuristic baseline is \emph{random}: we randomly select one of the options in case the question has multiple options (task 3 and 7), a number between 0 to 100 for questions having a numerical answer and a random entity present in the passage for questions having a text segment from the passage as the answer. 
In the \emph{majority} baseline, we select the most frequent answer for each task such as "Entailment" for NLI questions and similarly, the most frequent number for questions having numerical answer and the major entity present in the passage for questions having span based answer. 
As the task information is known, we include these baselines under task-specific baselines when discussing results.
\\ \\
\textbf{Zeroshot and Fewshot Baselines.}
We use GPT3 \cite{NEURIPS2020_1457c0d6} and the more recent GPT3-Instruct\footnote{newly released by OpenAI as part of the GPT3 finetuned series}. 
We have two types of few shot baseline (i) task specific and (ii) multi task. In case of task specific fewshot baseline, instances of the same task are used as in-context examples \cite{NEURIPS2020_1457c0d6} whereas in case of multitask few shot baseline, instances from all tasks are used to condition the model. 
Multitask fewshot is naturally a harder setting as it is task-agnostic. We use default parameters in GPT3 and GPT3-Instruct. In few-shot setting, we experiment after feeding as many examples as it can fit within the tokensize. For few shot experiments, we randomly select examples and averaged the results over 5 runs. 
\\ \\ 
\textbf{Fine-tuning Baselines.}
We first consider variations of the fine-tuning baselines in the context of our neuro-symbolic model, Ex-NumNet. 

We use it as bias-checking baseline -- to ensure that solving the benchmark correctly requires considering all of the information presented to it. 
To this end, we evaluate the performance of our model when finetuned only on the question (Q-only) or the context (C-only). 
% Several works have shown that some of the popular NLP datasets have annotation biases which are exploited by language models \cite{poliak2018hypothesis,gururangan2018annotation,bras2020adversarial,kaushik2018much}. So, we evaluate our benchmark using bias-checking baselines. We train Ex-NumNet model by heuristically removing the question completely and replacing it with the word "question" (Q-only baseline). We also do the same by removing context (C-only baseline). Context is different for different types such as passage in type 5 and type 6, premise in type 7, all sentences apart from question for type 8. In some minor cases, we split data manually where data could not be split automatically. 
Next, we present task-specific and multi-task baselines where Ex-NumNet is fine-tuned on individual tasks and the entire \numglue~benchmark respectively. 
With the goal of addressing the data imbalance across the tasks, we include an oversampling baseline that oversamples data from tasks with limited data so as to ensure that the model sees the same number of examples from each constituent task. 
% We also have a Data oversampling baseline where we try to tackle data imbalance by oversampling each type of data to the maximum size of all types.
% \ak{Here, introduce bias-checking, single task and multi-task and oversampling.}
% \\ \\ 
% \ak{In addition, we propose a new architectural modification to the numnetv2 by augmenting it with a memory. }
% \ak{End-to-end finetuning.}
% \textbf{Bias-checking Baselines }
% \textbf{Data Oversampling Baseline}
% We try to tackle data imbalance by oversampling each type of data to the maximum size of all types.
\\ \\
In addition, we propose a new architectural modification to Ex-NumNet. Noting that our baseline model Ex-NumNet does not take into account external knowledge, we create a new enhanced architecture in the form of a memory-augmented model that does Information Retrieval (IR) \cite{khot2019s} with respect to a knowledge base we create, \textit{MATH KB} to identify the needed knowledge. This is inspired by the observation that formula book and mathematical knowledge make the task easier for humans while solving math questions of various types. We then use this knowledge in the Ex-NumNet setting. Figure \ref{fig:knowledgemodel} illustrates our approach which leverages our newly created knowledge base \textit{MATH KB}. Conditional IR model is different from the regular IR model in the sense that, IR is performed only for questions of task 1 , 2 and 4, since they require external knowledge to get answered. More details about the model and the IR process can be found in supplementary material: Proposed Memory-Augmented Model (\ref{subsec:exnum} and \ref{subsec:mamodel}).
\\ \\
Finally, we discuss fine-tuning baselines in the context of end-to-end models, specifically GPT3.
We finetune the GPT3-13B model (for which the finetuning capability has been recently provided by OpenAI \footnote{https://beta.openai.com/docs/guides/fine-tuning}) in the multi-task setting i.e. the desired setting of the \numglue~benchmark.
% Because of restriction on number of finetuning experiments and token usage, we could only find the multitasking baseline. Adding IR and OS module to GPT3 will be part of future work.
\\ \\
\textbf{Human Baseline.}
Human baseline was calculated on 100 test set samples of each task (81 of Task 1) by averaging the scores of four annotators.

\section{Results and Discussion}
Table \ref{tab:main} shows the performance of various baseline models on the test set of our benchmark. 
Note that the performance of all baseline models is significantly lesser than the human baseline (Figure \ref{fig:teaser}). 
% Answering questions in eight different setting using a single model is indeed a challenging task. 
% has answered text when the answer is supposed to be a number. Similarly, it also has answered from span when it is supposed to select an option in MCQ setting.
% Table \ref{} shows t
We now discuss various insights based on these results. 
\\ \\
\textbf{Does the benchmark contain bias that a model can exploit?}
A challenging dataset requires the model to ideally consider all the information provided to it before arriving at an answer. 
To ensure that this is indeed the case, we perform ablations where only one portion of the input is provided i.e. either the question or the context. 
Both these ``bias-checking'' baselines perform poorly even in task-specific setting -- indicating that both the benchmark and constituent tasks are challenging.
% All bias-checking baselines do not perform well even with the help of a type oracle.
% We added type oracle to the heuristic baseline so that
% We create favorable condition for the models to exploit bias as much as possible using our top four baseline mentioned in Table. \ref{results}. 
% This shows that our dataset has very less bias.
\\ \\
\textbf{Which Tasks are Hard to Solve?}
Our results show that task 1 which requires numerical commonsense knowledge, is the hardest task to solve.
% Interestingly, the performance increases on providing the required knowledge by retrieving it from a cheat sheet.
% Our model which uses knowledge hunting significantly improves the performance on this category of questions.
Similarly, tasks 2, 4 and 8 appear to be comparatively harder from the rest. 
One pattern among these tasks is that all of them expect the answer to be numeric. 
Numeric answer requires accurate calculation. 
So, models might have difficulty in learning the task directly from data. 
This hypothesis is also justified from the \emph{slight} drop in human performance in these tasks..
\\ 
%\paragraph{Which question types are comparatively easier?}
On the other hand, task 7 has the best performance among all. 
Further, we see that performance on task 6 is slightly better than task 5 -- although both tasks are sourced from the same dataset, we observe that models answer span based questions better as compared to numeric answers. 
Relatively higher performance for task 3 suggests that models find it easier to answer in an MCQ setting.
\\ \\
\textbf{Does IR Help?}
Results show that knowledge help in improving performance of tasks 1, 2 and 4 -- where indeed, external knowledge like commonsense or domain-specific knowledge is needed in addition to arithmetic reasoning  to arrive at the correct answer.
However, task 3 is an exception to this trend and in fact registers a drop in the score when provided with (unnecessary) additional information; we find that this shortcoming is fixed when using conditional information retrieval (CIR) which in fact leads to the strongest baseline presented in this work.
% Not surprisingly, adding additional knowledge doesn't make any significant different to other tasks.
% It does not have significant difference for type 5, 6, 7, 8 which is as expected since questions of those types do not need external knowledge. This has been discussed in Section \ref{createdata}. However, seems like it acts as noise for type 3.
%This is because, most of the question types in this dataset do not need additional knowledge. So, knowledge acts as noise for them. 
% Conditional knowledge retrieval helps to mitigate the adverse effect and produces the best baseline.
\\ \\
\textbf{Does Oversampling help overcome data imbalance across tasks?}
Even though oversampling results in higher performance in certain tasks (in comparison with the multitask baseline), specifically the ones with smaller training data, it results in significant drop in performance in the other extreme, i.e tasks with bigger training data. Also, it never performs better than the Conditional IR module in multitask setting.

\subsection{Error Analysis}
% We find that, in some of the cases, the model fails to distinguish among the question types. For example, it gave a span based answer where a number was expected and vice versa.
We now present an analysis of the errors made by our baselines to indicate potential avenues for future research.
\\ \\
We analyze errors associated with 50 samples each of the 8 tasks and find that there are mainly 4 categories of error models make: (1) producing invalid output (e.g. answering text where the answer is supposed to be a number, answering a text different from the classes allowed in a classification problem), (2) copying a number from the question instead of calculating the answer, (3) incorrect calculation -- this can be due to multiple reasons including (i) using an incorrect operation e.g. subtraction in place of addition, (ii) incorrect parsing of numbers or (iii) incorrect knowledge of numerical commonsense facts. (4) producing redundant text after producing correct answer. 
Based on error distribution in Table \ref{tab:error}, we observe that the majority of errors come from incorrect calculation. 
Further, GPT3 is better than Ex NumNet+v2 in producing valid outputs, but it produces more redundant text.
\\ \\
\textbf{Future Directions: Bigger model, more data or $\dots$?}
Table \ref{tab:main} shows that fine-tuned GPT3-13B outperforms other baselines on task 1, 2 and 3.
Recall that these tasks require external knowledge and perhaps, this is the reason why GPT3, already pre-trained on a diverse web-scale text corpus has an edge over other baselines on these tasks.
% This may be because GPT3 being a larger end-to-end model pre-trained on a huge diverse corpus might have already acquired some of the required knowledge. 
In case of the smaller Ex-NumNet, it is interesting that multitask baselines are higher than the single task baselines by 3.4\% on average and that information retrieval helps in tasks that require external knowledge. Also notice that, GPT-3 is better on smaller datasets and NumNet is better on large datasets. This may indicate that GPT-3 is a better few-shot learner but not necessarily a better many-shot learner.
% More experiment needs to be performed to verify this hypothesis.
This non-overlapping performance of GPT-3 and Ex-numnet, end-to-end and neuro-symbolic models respectively, indicates that a potential future direction for research is to combine the best of both the models.
% So adding symbolic and IR module to GPT3 may be able to improve baselines. Alternatively, extracting required knowledge from GPT3 and using it in Ex.NumNet+v2 can be another interesting direction from accessibility and sustainability \cite{schwartz2020green} point of view.
\begin{table}[t]
    \small
    \centering
    % \resizebox{\columnwidth}{!}{%
    \begin{tabular}{p{2.8cm}p{2cm}p{1.5cm}}
    \toprule
        Error & Ex-NumNet & GPT3 \\ 
\midrule
         Invalid output &  16 \% & 7\% \\
         Copy number &  5 \% &  3\%  \\
        Incorrect calculation &  71 \% &  56\%  \\
        Redundant text &  8 \% &  34\% \\
\bottomrule

    \end{tabular}
    % }
    \caption{Error analysis for the best Ex-NumNet Multitask+CIR and GPT3 Task-specific model}
    \label{tab:error}
\end{table}
 \section{Conclusion}
 We propose \numglue, a multi-task benchmark to test for arithmetic understanding. 
 Our benchmark consists of eight tasks including four new ones.
 While some of the tasks require external knowledge like commonsense or domain-specific information in addition to arithmetic reasoning, some are self-contained e.g. arithmetic word problems.
 Further, we demonstrate that our benchmark is far from being solved -- with state-of-the-art large scale models achieving considerably lower performance than humans.
 This indicates that current AI systems are incapable of performing simple arithmetic reasoning in a general setting -- indicating a fundamental hurdle towards AI systems that understand complex mathematical concepts like differential equations or combinatorics.
 Finally, we present various baselines including a novel architecture (memory augmented Ex-NumNet) that demonstrate the advantages of various modeling choices (e.g. end-to-end vs neuro-symbolic models).
 Specifically, we show that training in the multi-task setting leads to meaningful sharing of knowledge across tasks as evidenced by an average gain of 3.4\% on tasks compared to task-specific modeling.
% \cha{A potential future work is to increase the size of Task 1-4, specifically Task 1 where data creation is constrained by lack of numerical commonsense facts e.g. human has 2 hands. Unit conversion and other facts involving numbers contain the risk of requiring knowledge beyond commonsense. Adding more complex reasoning problems to \numglue can be another future work.}
Finally, we hope that our benchmark not only leads to AI systems that are capable of performing simple arithmetic reasoning in a fairly general setting but also results in progress towards more complex mathematical reasoning capability.  
 \section*{Acknowledgements}
We thank OpenAI for providing academic access to the GPT3 API, the Aristo team at AI2 for helpful input, the Beaker team for their support with experiments and the anonymous reviewers for their insightful feedback. The support of DARPA SAIL-ON,  DARPA CHESS program is
gratefully acknowledged.

\section*{Ethical Considerations} 
We have verified that all licenses of source datasets used in this paper allow for their use, modification, and redistribution in a research context. The dataset will be distributed in a manner similar to SuperGLUE ~\cite{wang2019superglue} i.e. give full credit assignment to the original data and task creators. 

\bibliography{anthology,custom}
\bibliographystyle{acl_natbib}
\clearpage
\appendix

% \section{Example Appendix}
\label{sec:appendix}
\section{Supplemental Material}
\label{sec:supplemental}
\subsection{\numglue~vs Other Datasets:}
\begin{figure*}[t]
\centering
\begin{minipage}{\textwidth}
   \centering
   \includegraphics[width=0.78\textwidth]{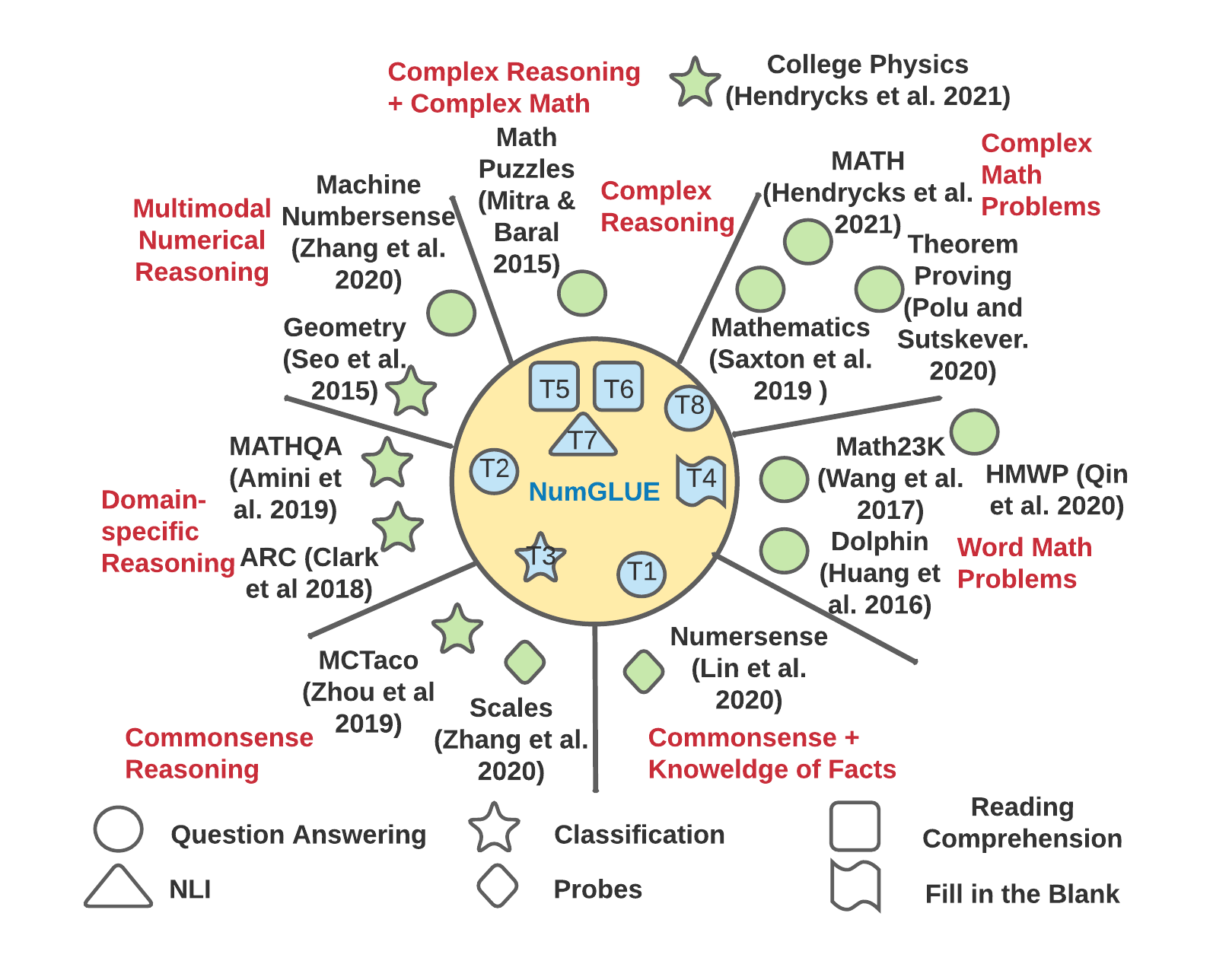}
%   \captionof{figure}{A figure}
%   \label{fig:test1}
\caption{Our dataset \numglue~(center in the yellow circle) has been positioned with respect to existing datasets. T1-T8 represents 8 tasks. Note that, \numglue~contains the feature of being format invariant unlike other datasets. Position of datasets within clusters is done based on their semantic category, for example T1 Numerical Commonsense QA is closer to the cluster of Commonsense Reasoning + Knowledge of Facts; its position reflects the same }
\label{fig:map}
\end{minipage}%
\end{figure*}

As figure \ref{fig:map} shows, we select each task from one of the clusters of numerical reasoning datasets (except the multi-model reasoning cluster since we wanted to limit our dataset to text only).
\subsection{Construction of \numglue~:}
\begin{figure*}[!htb]
\centering
\begin{minipage}{\textwidth}
   \centering
   \includegraphics[width=\linewidth,height=8cm,keepaspectratio]{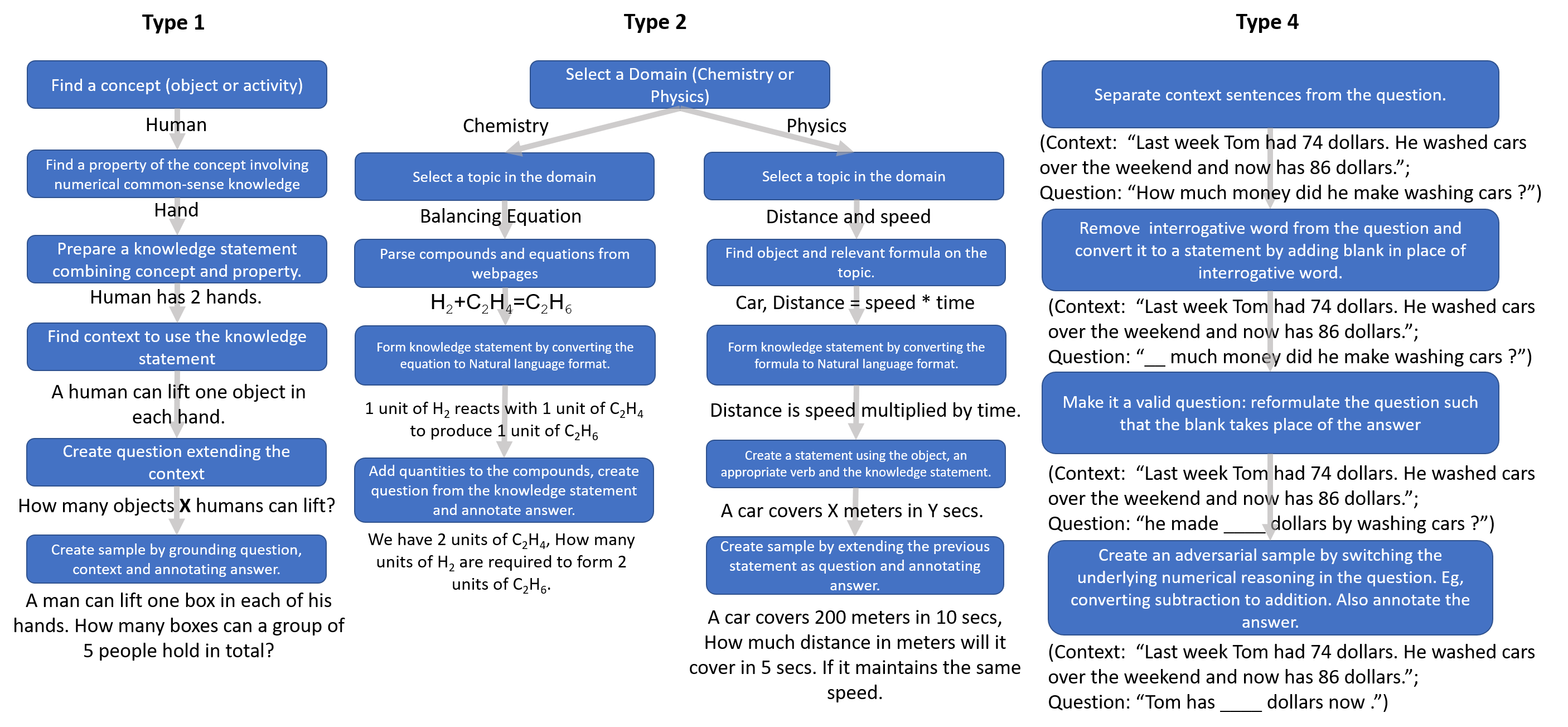}
%   \captionof{figure}{A figure}
%   \label{fig:test1}
\caption{Step by step data creation process for task 1, 2 and 4 questions}
\label{fig:Flow1}
\end{minipage}%
\end{figure*}

\begin{figure}[!htb]
\centering
\begin{minipage}{.5\textwidth}
   \centering
   \includegraphics[width=\linewidth,height=8cm,keepaspectratio]{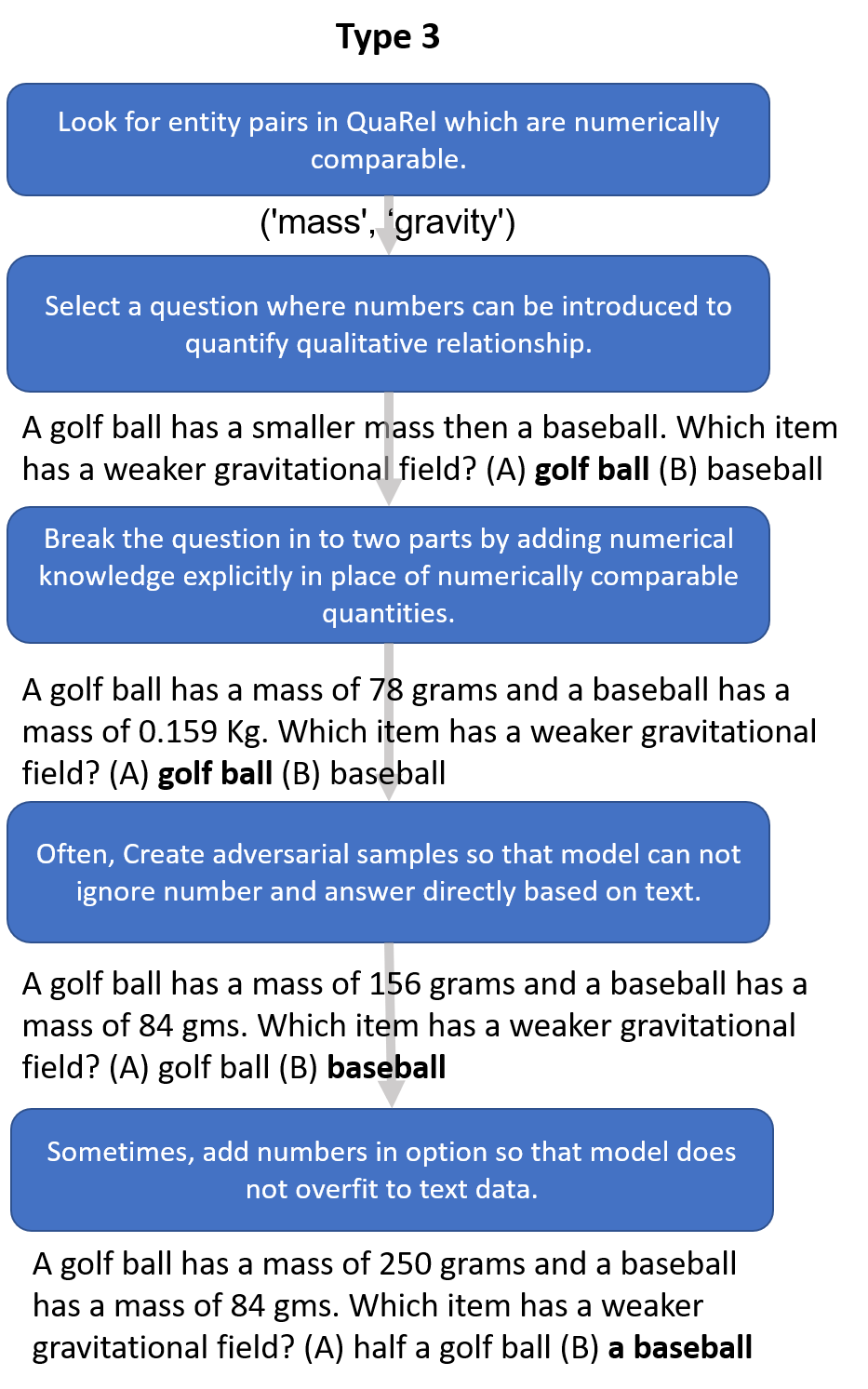}
%   \captionof{figure}{A figure}
%   \label{fig:test1}
\caption{Step by step data creation process for task 3 questions}
\label{fig:Flow3}
\end{minipage}%
\end{figure}

Figure \ref{fig:Flow1} and \ref{fig:Flow3} illustrate detailed data creation process for task 1, task 2, task 3 and task 4 questions with the help of an example for each task. We follow the same procedure for creating other examples within the task.
% \begin{figure}
% \centering
% \begin{minipage}{.5\textwidth}
%   \centering
%   \includegraphics[width=\linewidth,height=8cm,keepaspectratio]{NumericalReasoningType2Flow.png}
% %   \captionof{figure}{A figure}
% %   \label{fig:test1}
% \caption{Flow 2}
% \label{Flow2}
% \end{minipage}%
% \end{figure}

% \begin{figure}
% \centering
% \begin{minipage}{.5\textwidth}
%   \centering
%   \includegraphics[width=\linewidth,height=8cm,keepaspectratio]{NumericalReasoningType4Flow.png}
% %   \captionof{figure}{A figure}
% %   \label{fig:test1}
% \caption{Flow 4}
% \label{Flow4}
% \end{minipage}%
% \end{figure}

\subsection{GPT3-Instruct's Response}
\label{subsec:GPT3-Instruct}
\begin{figure*}
\includegraphics[width=\textwidth]{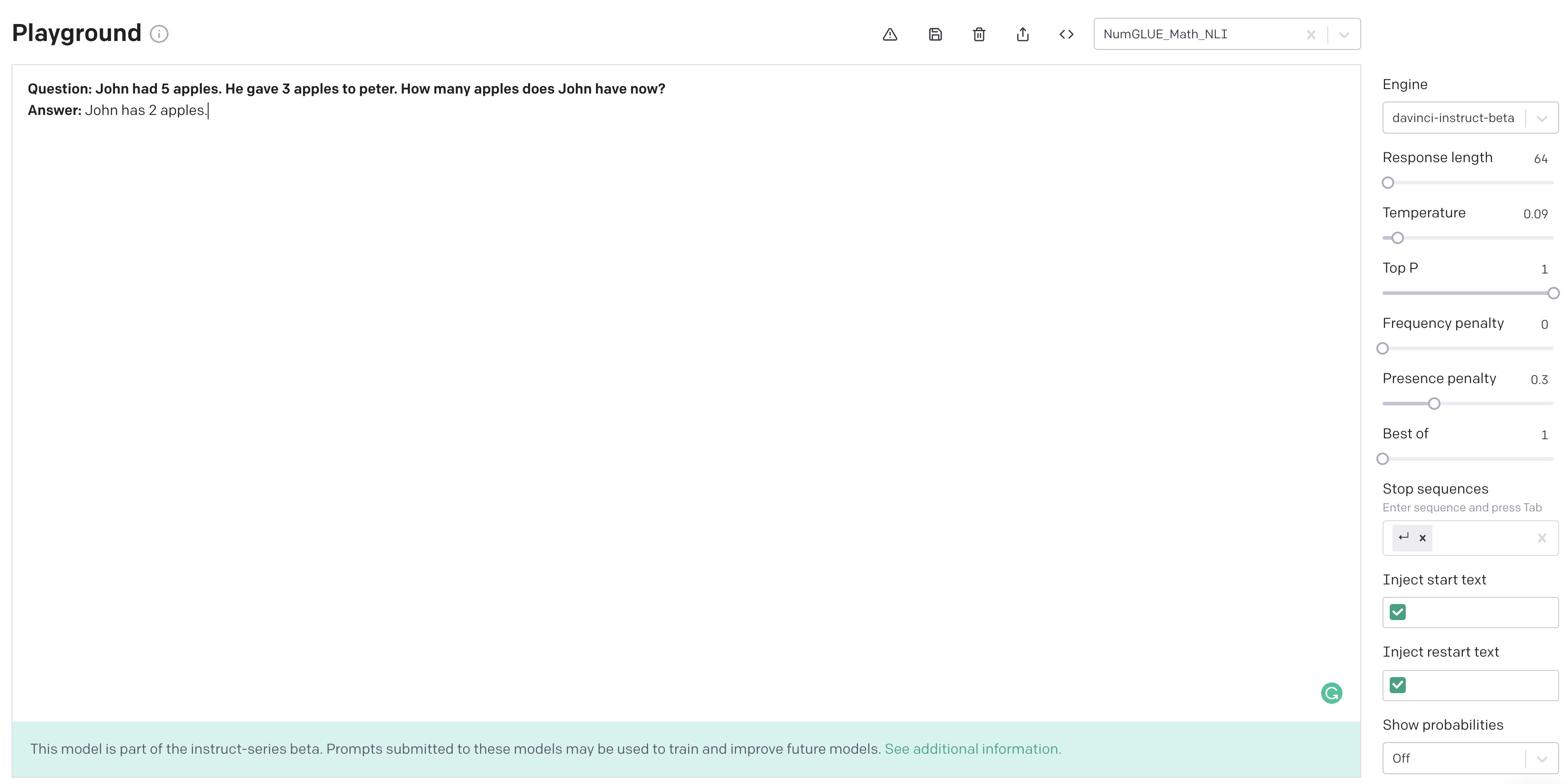}
  \centering
  \caption{GPT3-Instruct's response to a simple numerical reasoning question.}
  \label{fig:r1}
\end{figure*}

\begin{figure*}
\includegraphics[width=\textwidth]{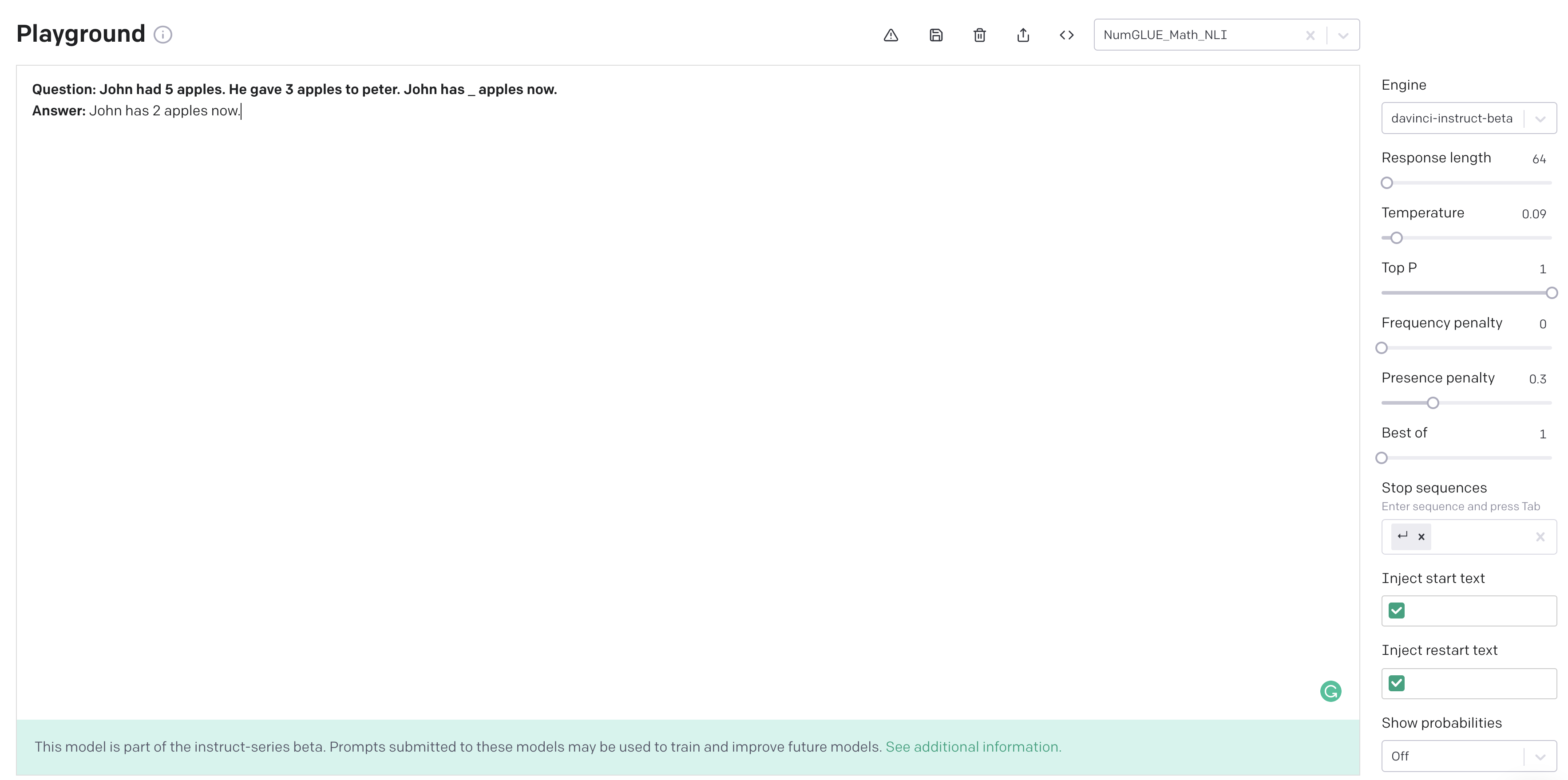}
  \centering
  \caption{GPT3-Instruct's response to a simple numerical reasoning question expressed in fill in the blanks format.}
  \label{fig:r2}
\end{figure*}

\begin{figure*}
\includegraphics[width=\textwidth]{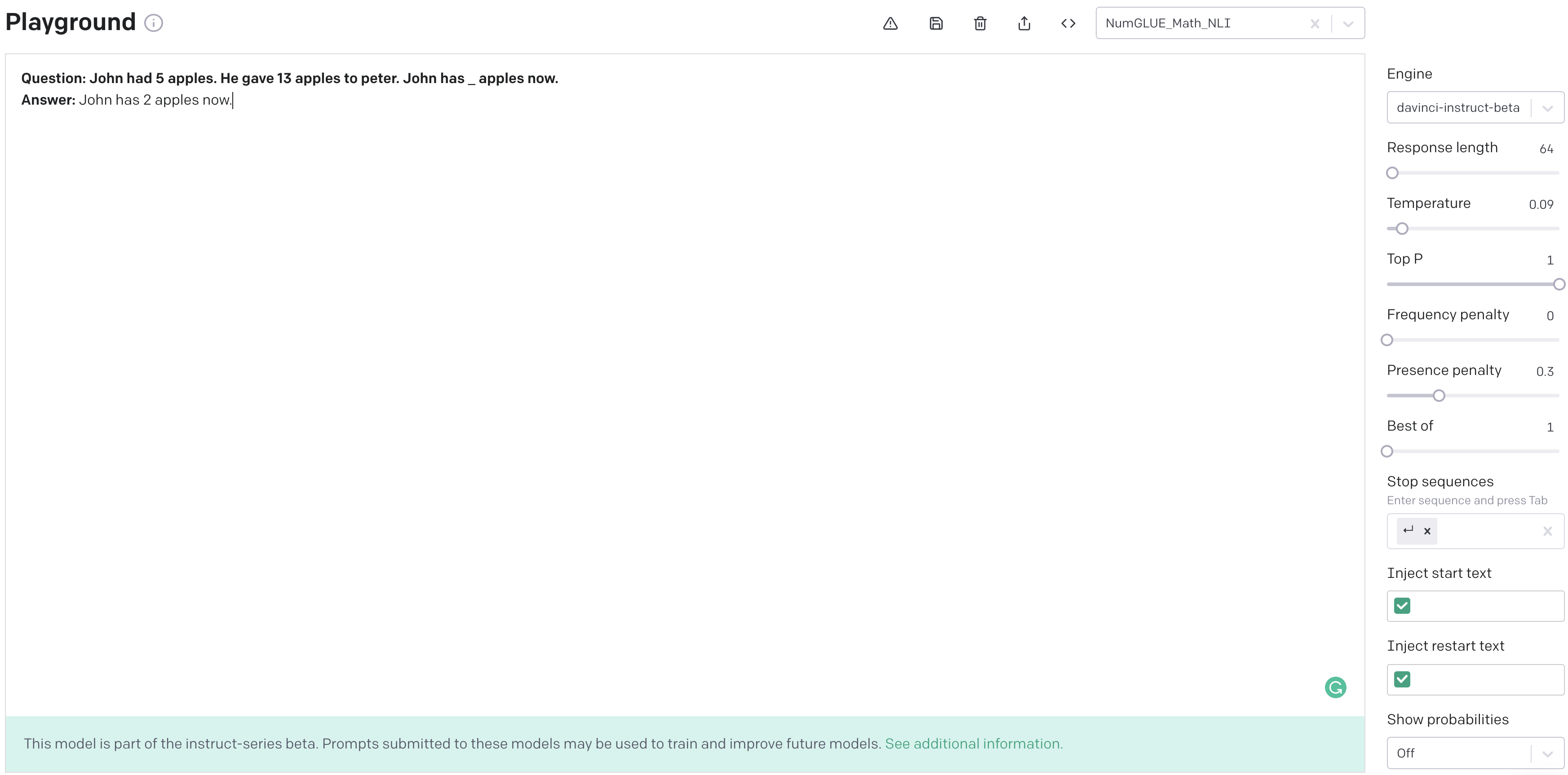}
  \centering
  \caption{GPT3-Instruct's response to a simple numerical reasoning question expressed in fill in the blanks format where numbers are changed.}
  \label{fig:r3}
\end{figure*}

\begin{figure*}
\includegraphics[width=\textwidth]{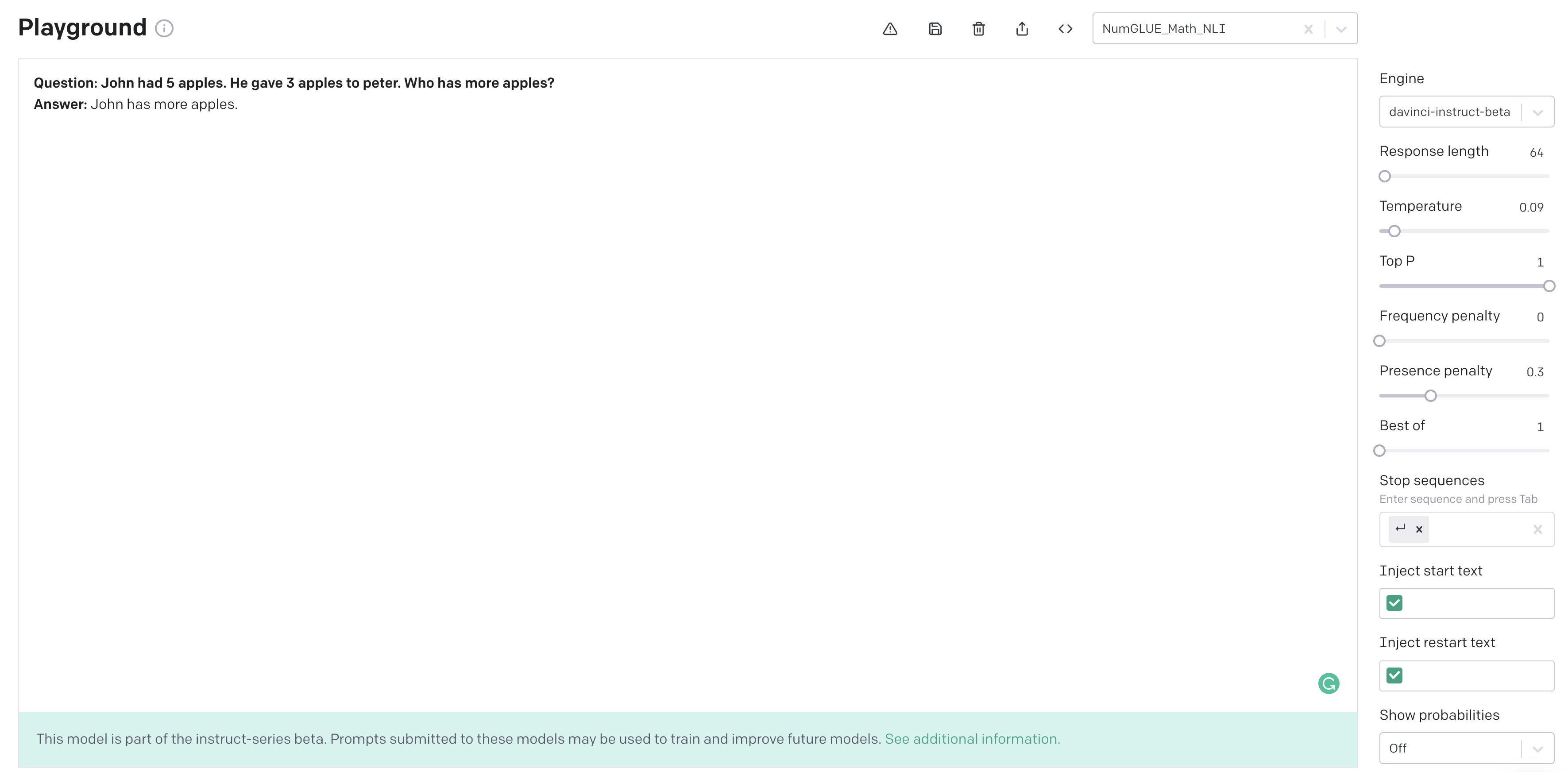}
  \centering
  \caption{GPT3-Instruct's response to a simple numerical reasoning question expressed in comparison format.}
  \label{fig:r4}
\end{figure*}

\begin{figure*}
\includegraphics[width=\textwidth]{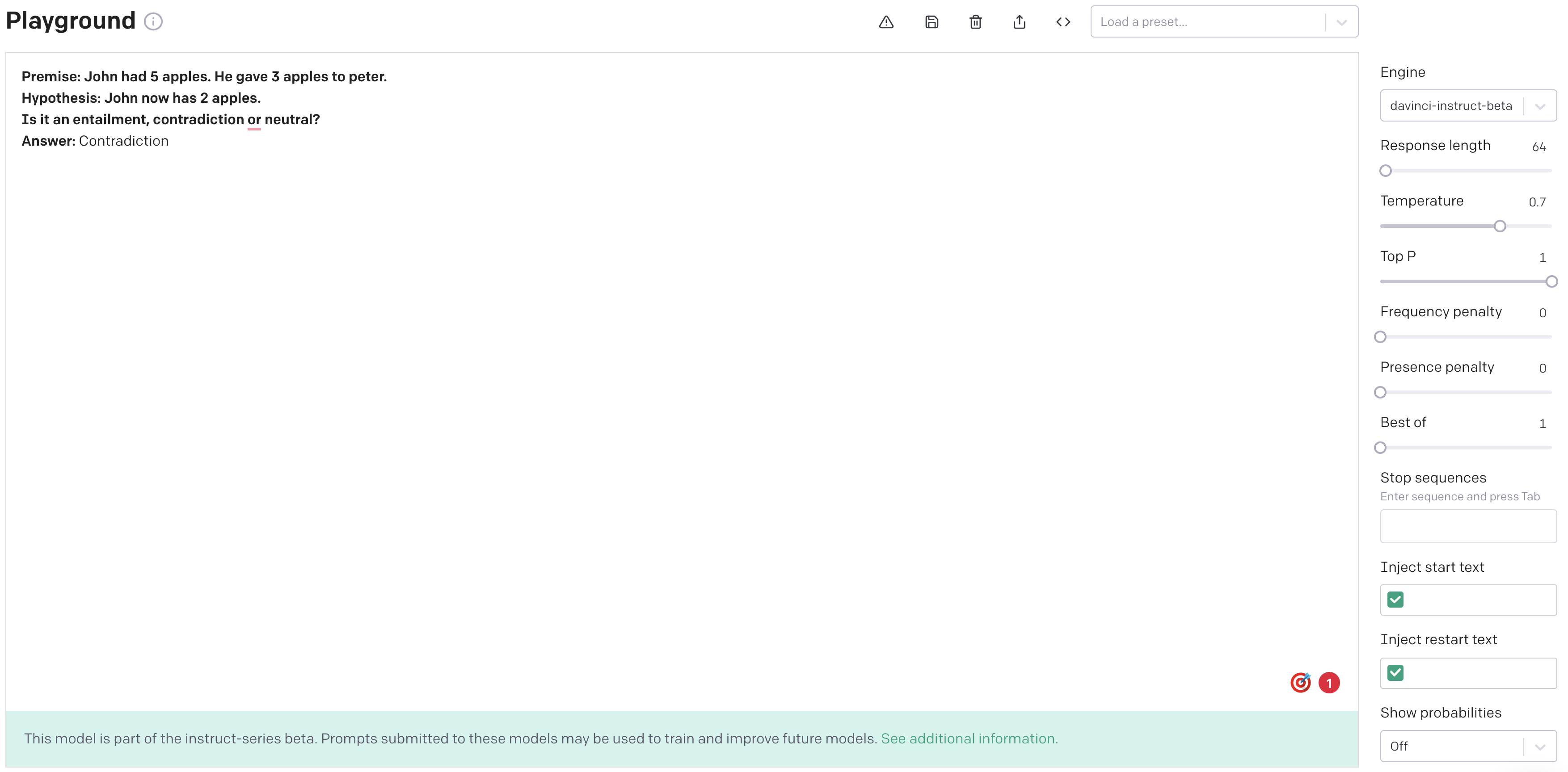}
  \centering
  \caption{GPT3-Instruct's response to a simple numerical reasoning question expressed in NLI format.}
  \label{fig:r5}
\end{figure*}

We used GPT3-Instruct on various forms of a simple arithmetic question. An expert did tuning of various parameteres such as temperature, stop condition, presence penalty, engine, maximum token size. However, GPT3-Instruct still could not solve the basic aritmetic questions reliabily (Figures \ref{fig:r1}-\ref{fig:r5}).
% \subsection{MathKB:}
\subsection{Data Quality Analysis of NumGLUE}
\begin{figure*}[t]
    \centering
    \begin{subfigure}{0.48\textwidth}
      \centering
      \includegraphics[width=1\linewidth]{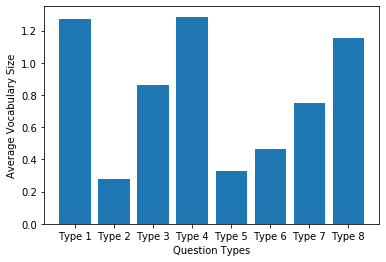}
      \caption{Average vocabulary represents the average number of unique words across various tasks. On an average, novel datasets (task 1-4) have higher vocabulary.}
      \label{fig:vocab}
    \end{subfigure}
    \hspace{0.01cm}
    \begin{subfigure}{0.48\textwidth}
      \centering
      \includegraphics[width=1\linewidth]{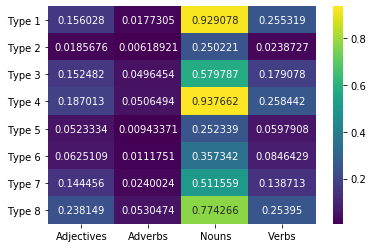}
      \caption{ Average number of unique Part of Speech (POS) tags is higher for task 1 and task 4 in the novel datasets in contrast to other tasks.}
      \label{fig:pos}
    \end{subfigure}
    \begin{subfigure}{0.24\textwidth}
      \centering
      \includegraphics[width=1\linewidth]{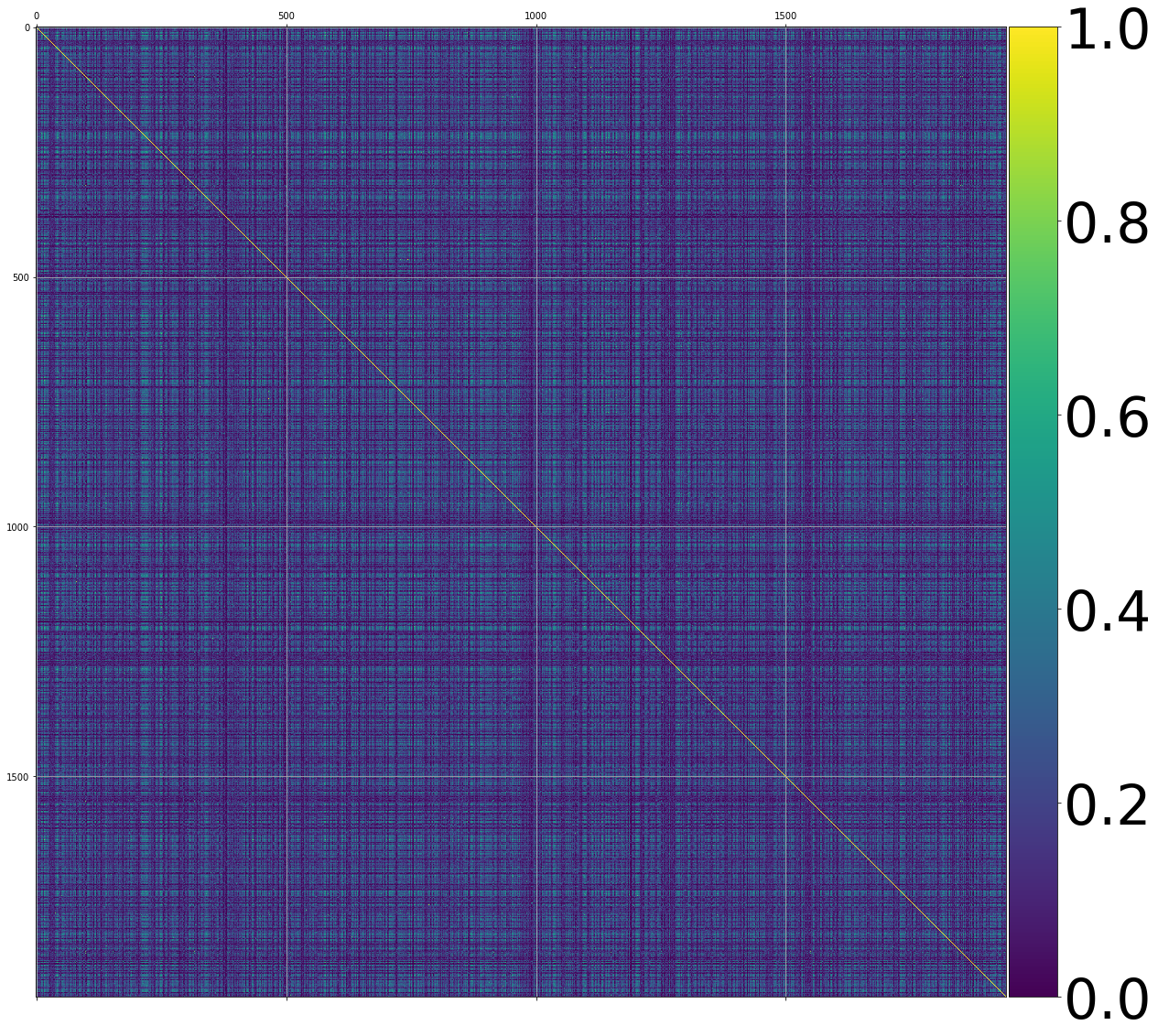}
      \caption{STS plot for the QuaReL dataset shows significant repetition across samples}
      \label{fig:sts1}
    \end{subfigure}
    \hspace{0.01cm}
    \begin{subfigure}{0.24\textwidth}
      \centering
      \includegraphics[width=1\linewidth]{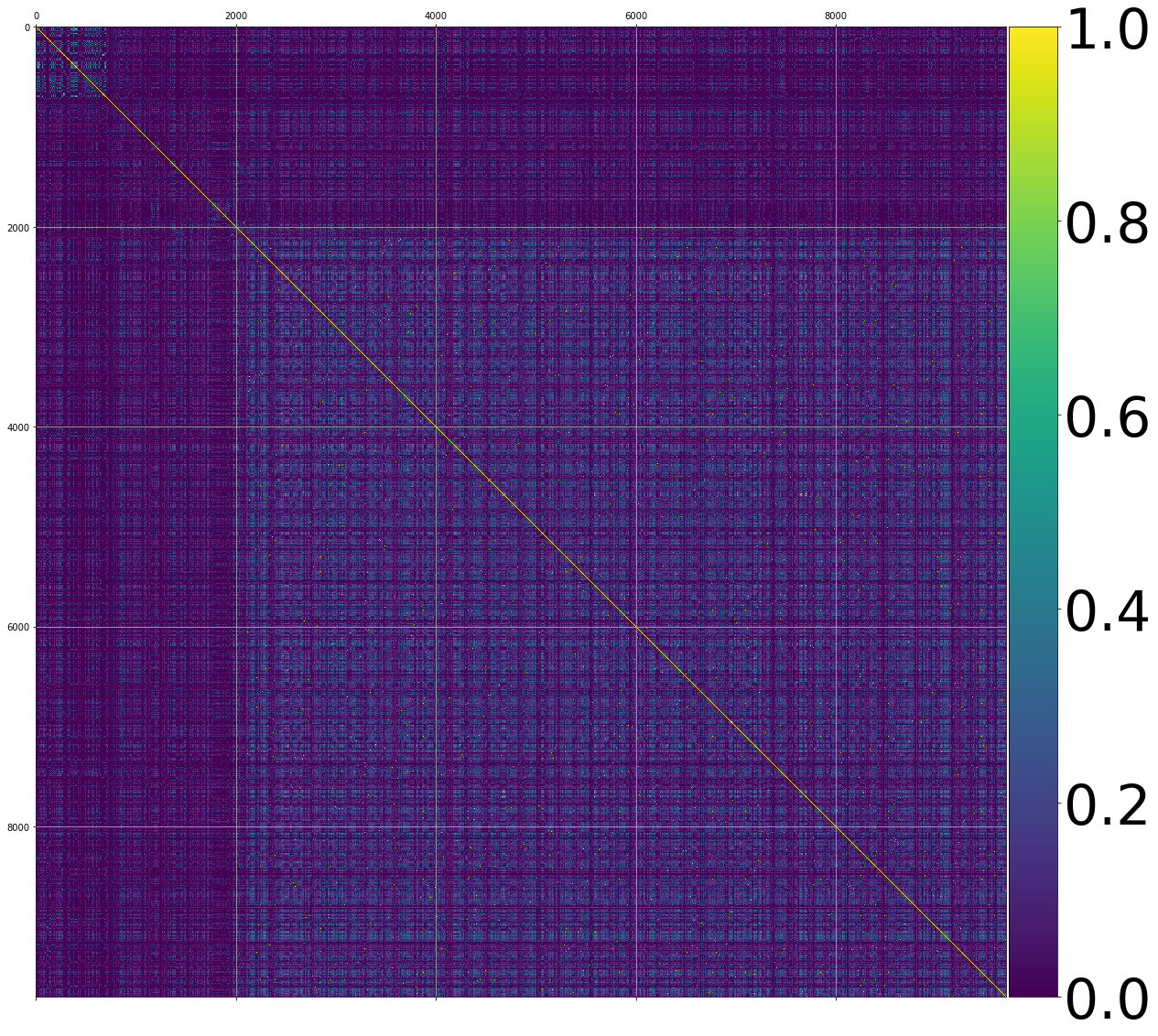}
      \caption{STS plot for the EQUATE dataset shows considerable repetition across samples.}
      \label{fig:sts2}
    \end{subfigure}
    \hspace{0.01cm}
    \begin{subfigure}{0.24\textwidth}
      \centering
      \includegraphics[width=1\linewidth]{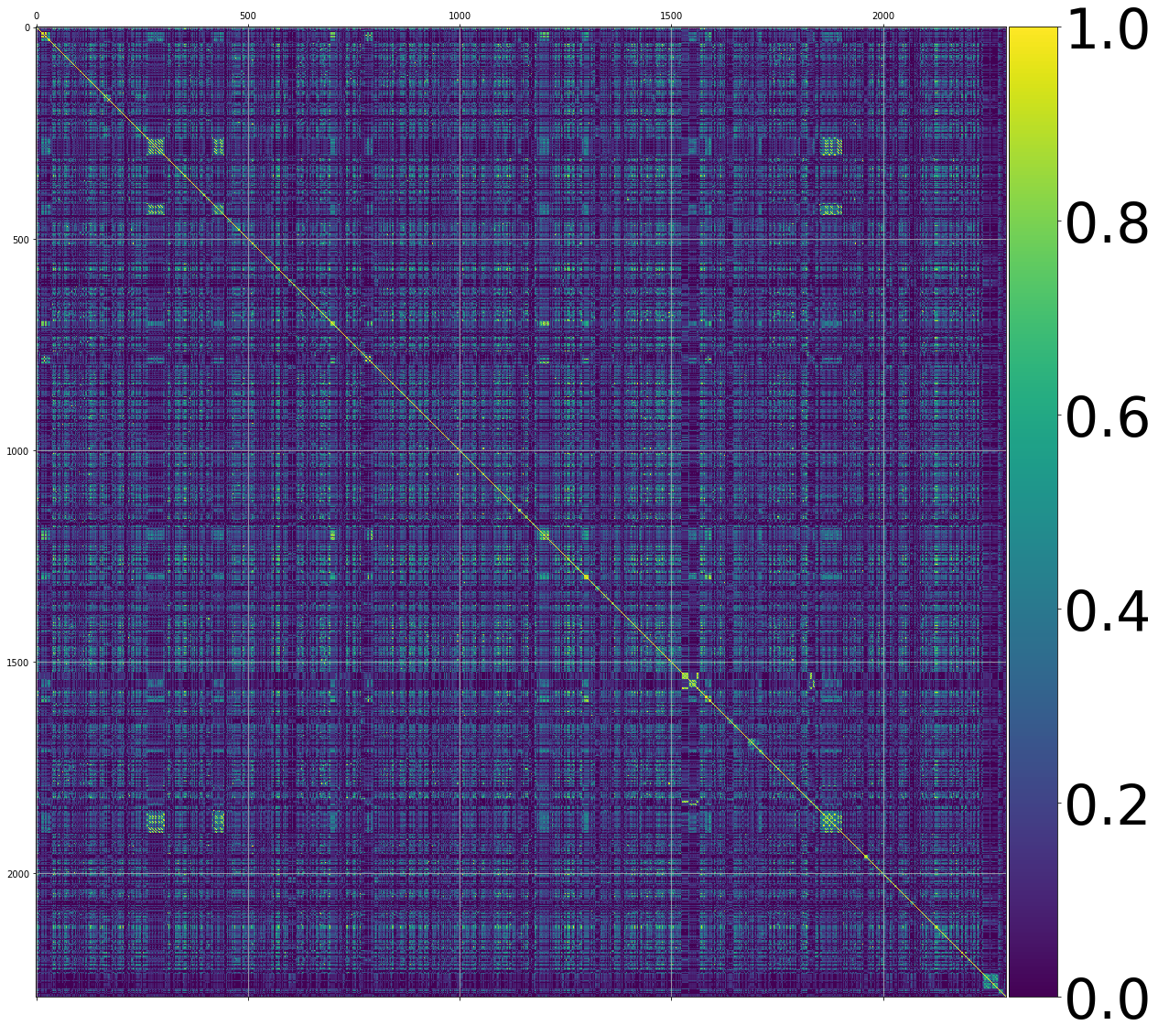}
      \caption{STS plot for the DROP dataset shows repetitions for most part of the data.}
      \label{fig:sts3}
    \end{subfigure}
    \hspace{0.01cm}
    \begin{subfigure}{0.24\textwidth}
      \centering
      \includegraphics[width=1\linewidth]{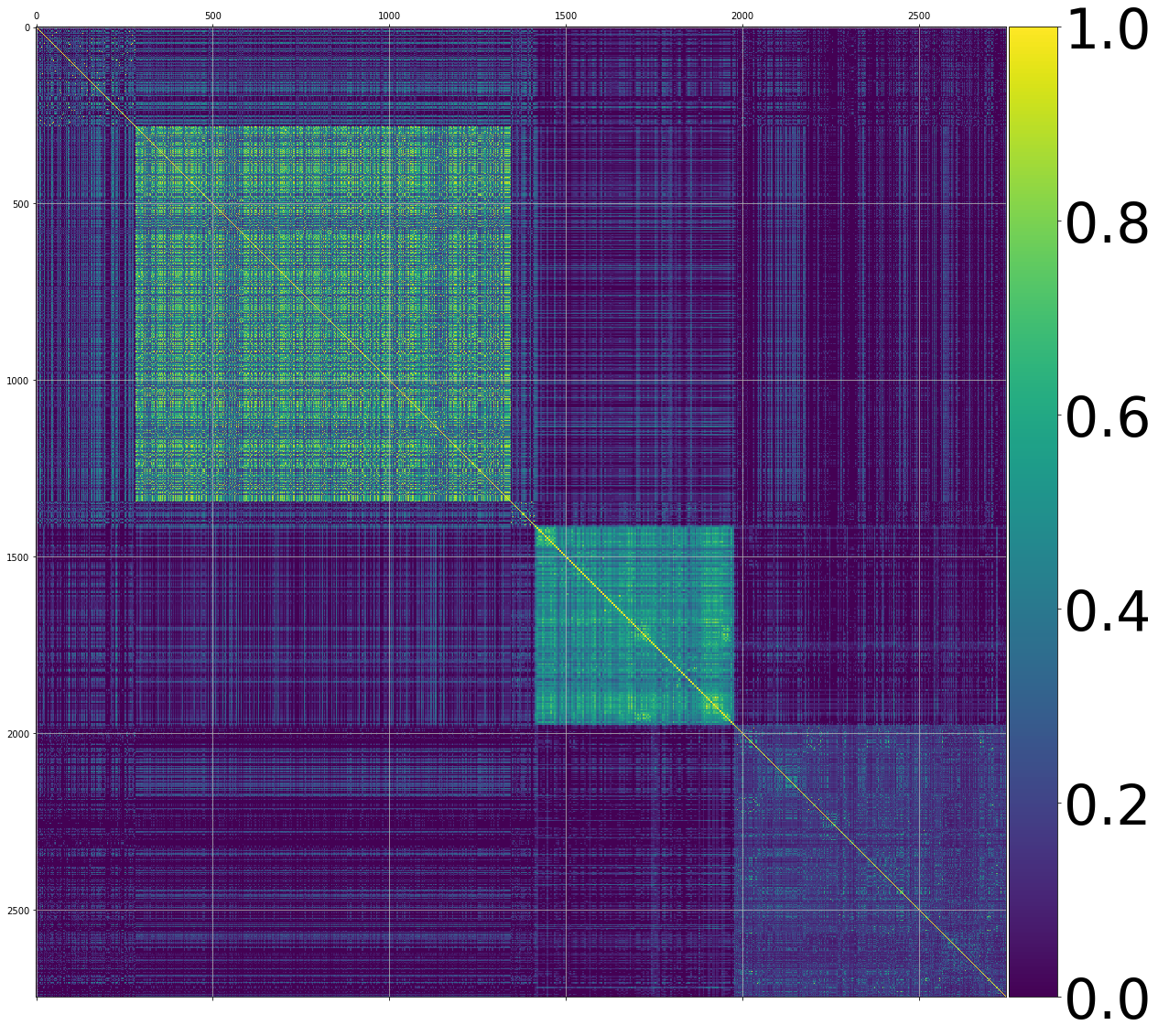}
      \caption{STS plot for the novel datasets show relatively lower repetition than other datasets}
      \label{fig:sts4}
    \end{subfigure}
     \caption{Data quality analysis of \numglue~across various tasks of data. On an average, novel datasets have higher quality than the others since they have higher average vocabulary, higher average POS tag numbers and lower Semantic Textual Similarity (STS) among each other. X-axis and Y-axis represents samples ordered in the same way, an ideal high quality dataset would have a bright line in the diagonal and rest of the places it should be dark signifying lower repetition across instances.}
\end{figure*}

In this section, we discuss various linguistic and statistical properties of our benchmark; ones that we believe result in the quality, diversity and challenging nature\cha{~\cite{gururangan2018annotation, mishra2020dqi, mishra-sachdeva-2020-need, swayamdipta2020dataset, mishra2020our, arunkumar2020real}} of the proposed \numglue~benchmark.
\\ \\
\textbf{Vocabulary Size.} First, we calculate vocabulary size of each task by finding the number of unique words across all questions. Since our dataset is unbalanced in terms of question task, we find the average vocabulary size by dividing vocabulary size with number of data in that task. 
% Figure \ref{VocabDistribution} illustrates average vocabulary size of each question type. 
% Average vocabulary size of Type 1 is 1.2 which implies that if there are a total of 100 questions in type 1 then there are 120 unique words across all the questions in type 1. 
\\ \\
\noindent \textit{Which Data has Higher Average Vocabulary?} As illustrated in Figure \ref{fig:vocab}, most of the tasks belonging to the novel dataset category have relatively better average vocabulary size. This implies 
questions in those tasks have less repetitiveness. 
Furthermore, we expand our vocabulary analysis to understand Figure \ref{fig:vocab} better. We dive deep to analyze different parts of speech. Figure \ref{fig:pos} summarises our analysis. Most of the novel datasets have more average number of nouns, verbs and adjectives implying there are more varieties of entities, actions and attributes. This further means that datasets belonging to the novel category are more diverse in nature.
\\ \\
\noindent \textbf{Sentence Similarity Analysis}
We extend our analysis to reinforce our inference from the word vocabulary analysis. We find Semantic Textual Similarity (STS) of a sentence with every other sentence. 
% Figure \ref{similarityChart_1} illustrate this for various datasets.
\\ \\
\textit{Which Data Consists of Most Dissimilar Sentences?}
As depicted by Figure \ref{fig:sts1}-\ref{fig:sts4}, most questions in QuaRel have high similarity value with other questions indicating the repetitiveness of data. Same is true for majority of EQUATE data. DROP also has high similarity among questions. 
%We also analyze this similarity metric for our dataset and find that similarity among questions is significantly less. 
However, similarity among questions in our dataset is significantly less.
Some similarity boxes can be seen in the chart. They are mostly due to task 2 data, and partly due to task 3 data. Lesser similarity implies that our dataset is far less repetitive than others. Also, the repetition in our dataset is sparse and is not equally distributed among the whole dataset unlike others. This way, our dataset is more diverse.
\\ \\
\noindent Note that question in Task 2 have lower vocabulary and further, a higher similarity as well.
As a small set of chemistry and physics principles are used to generate questions, the result is a fairly templated or uniform-looking dataset -- leading to the observed reversal of trends in this particular task. 
\\ \\
\subsection{Ex-NumNet}
\label{subsec:exnum}

\begin{figure*}
\includegraphics[width=\linewidth]{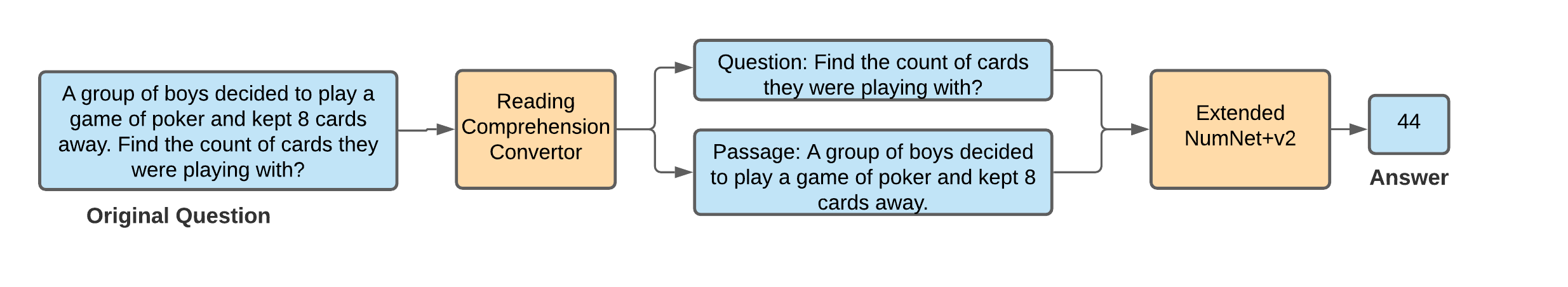}
  \caption{Architecture of Ex-NumNet}
\label{modl}
\end{figure*}

\begin{figure*}
\includegraphics[width=\linewidth]{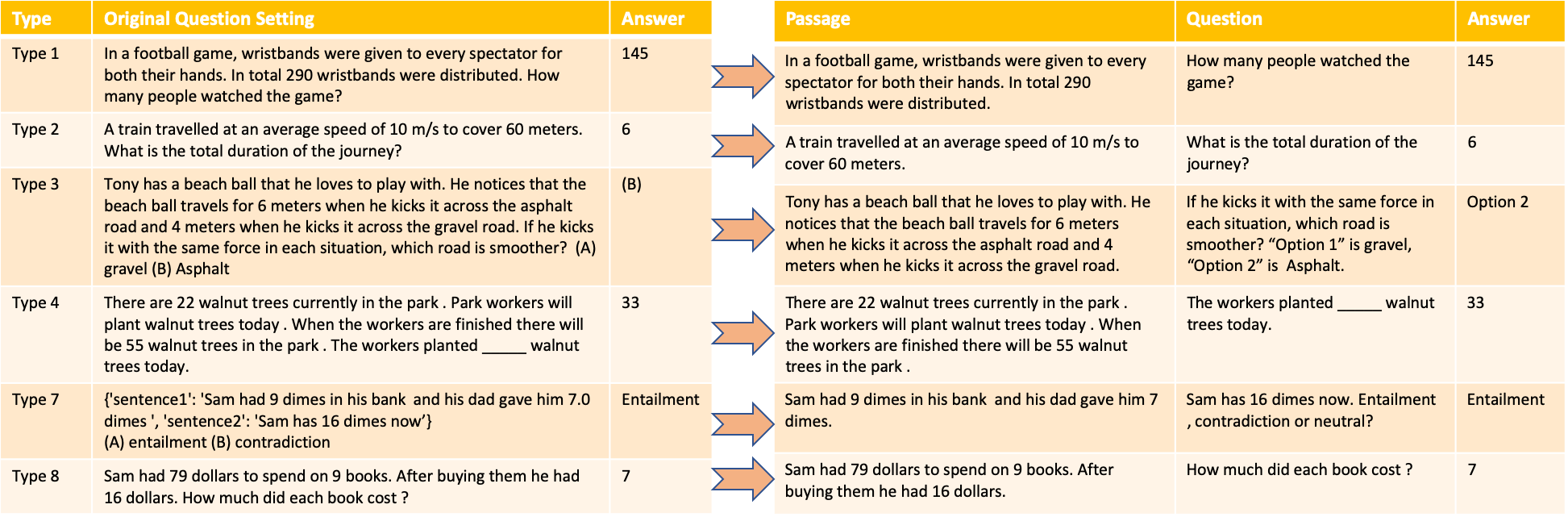}
  \caption{Conversion of various tasks to reading comprehension format}
\label{motiv}
\end{figure*}

Figure \ref{modl} illustrates our baseline model: Ex-NumNet. This contains a Reading Comprehension Converter module which converts each task of question to reading comprehension format. Figure \ref{motiv} illustrates various examples of how each task of questions get converted to the reading comprehension format. We add a task converter module to detect task of a question. We design task converter heuristically based on the features associated with questions (e.g. NLI contains "Sentence 1" and "Sentence 2" whereas completion contains a blank). We convert each of the tasks to RC format. For NLI questions, we use the premise sentence as passage, hypothesis as the question and append the string “Entailment, contradiction or neutral?” to the question so that it has a span based answer. For other questions, we tokenize the question string into its constituent sentences and use a heuristic approach to split the question string into passage and question. Furthermore, for option based questions, we append all the options at the end of the question.
\subsection{Proposed Memory-Augmented Model}
\label{subsec:mamodel}
Figure \ref{modl} illustrates our baseline model Ex-NumNet. We add an IR mechanism as described in Algorithm 1 and illustrated in Figure 3 of the main paper. As mentioned in the `Baselines' subsection (Experiments section) of the main paper, we convert each task to RC format in our baseline and append the knowledge retrieved using IR from \textit{MATH KB} at the end of the passage. In our experiments, we use the following hyperparameters in the IR process: $Z=50$, $v=10$, $th=0.75$ and $b=0.1$. 

\textbf{Formalization} Let $D$ represents dataset, $s$ represents sample, $K$ represent the \textit{MATH KB}, $v$ represents the number of knowledge statements retrieved for each sample, $th$ is the cut off STS (Semantic Textual Similarity) value above which knowledge statements are treated redundant and removed, $b$ is the reduction we do iteratively on $th$ until $v$ statements remain.

We create a knowledge base, \textit{MATH KB} by accumulating all tasks of external knowledge which are needed to solve questions of various tasks (e.g. human has 2 hands, cow has 4 legs, there are 24 hours in a day \etc.). We also add math formulae required to solve questions in our benchmark (e.g. the formula of speed in terms of distance and time). We add alll these in the form of plain text separated by new line.  We use Elasticsearch to retrieve relevant knowledge sentences. We further filter them using a heuristic threshold of relevance. We append this knowledge in the beginning of the passage so that continuity is not broken between passage and question. Figure 3 of the main paper illustrates our approach.\\

\begin{algorithm}[!htb]
\small
\SetAlgoLined
\KwInput{Dataset $D$, MATH KB $K$ \textbf{Hyper-Parameters}: $Z$, $v$, $th$, $b$}
\KwOutput{$v$ Knowledge sentences }
 \ForAll{$s \in D$}{
 Concat Question and Answer  \;
 Generate Query by retaining only verbs, adjectives and adverbs\;
 \ForAll{$j \in K$}{
 Create Index using Elastic Search \;
 Retrieve top Z sentences from MATH KB.
 }
 \While{size(Z)$>$v}{
 \ForAll{$k \in Z$}{ 
 \ForAll{$u \in k-1$}{
 \If{STS(Z(u),Z(k))$>$th}{
 Delete k\;}
 }
 }
 th=th-b\;
 }
  }
 \caption{Our Information Retrieval Approach}
 \label{algo:one}
% can also be prediction accuracies of RAFLite
\end{algorithm}
% \begin{figure*}
% \includegraphics[width=\linewidth]{FullMODEL5.png}
%   \caption{Overview of our baseline model Extended NumNet+v2 depicting the various questions types and their answers.}
% \label{fig:motivation16}
% \end{figure*}
%\begin{figure}
% \includegraphics[width=\linewidth]{FullMODEL5.png}
%   \centering
%   \fbox{\rule[-.5cm]{0cm}{4cm} \rule[-.5cm]{4cm}{0cm}}{FullMODEL5.png}
%   \caption{Sample figure caption.}
% \end{figure}
% \subsection{MathKB:}
\subsection{Hyper Parameters Used}
All the experiments were ran with the following hyper parameters, batch size was kept at 16 where as the eval batch size was 5. The maximum number of epoch ran for the experiments were 5 with the warm-up kept at 0.06. The learning rate used was 1.5e-5 and the weight decay was 0.01.

All above hyper parameters were selected using a grid search; we kept rest of the hyper parameters unaltered. All the experiments were performed on "TeslaV100-SXM2-16GB", with which the model takes 24hrs to train on nearly 100k samples.

\subsection{Additional Examples}
We provide additional examples of task 1, 2, 3 and 4 questions here to better illustrate the novel datasets we have created as part of our \numglue.

\begin{table}[h]
    \centering
    \resizebox{\linewidth}{!}{
    \begin{tabular}{ |l|l|r| }
    \hline
    % \toprule
        \textbf{Question} &
        \textbf{Knowledge Required} &
        \textbf{Answer}\\
    % \midrule
    \hline
    \hline
    %Jonathan and Levi are playing a game of chess, they have lost 4 and 1 pawns respectively.  Find the total number of pawns left in the game & In the game of chess, each player has 8 pawns initially. & 11 \\
    %\hline
    %Benjamin was leading a new project at work. So, he worked 7 hours every day for 2 weeks to finish the project. For how many hours did Benjamin work in this duration?
    %&
    %A week has 7 days
    %&
    %&98 \\
    %
    \multicolumn{1}{|p{3.4cm}|} {Ella and Lily are playing a game that requires 10 die. Find out the total number of faces in 10 die.}& \multicolumn{1}{|p{2.1cm}|}{A die has 6 faces}& 60 \\
    \hline
    % \midrule
    \multicolumn{1}{|p{3.4cm}|}{Jacob and Lillian are running a km long race. Jacob finished the race when Lillian was 190 meters from the finish line. How many meters did Lillian cover till that time?}& \multicolumn{1}{|p{2.1cm}|}{ 1000 meters make a km } & 810\\
    \hline
    % \midrule
    \multicolumn{1}{|p{3.4cm}|}{A man can lift one box in each of his hands. How many boxes can a group of 5 people hold in total?}& \multicolumn{1}{|p{2.1cm}|}{ A human being has 2 hands } & 10 \\
    
    %line
    % \midrule
    
    \hline
    % \bottomrule
    \end{tabular}
    }
    \caption{Example questions where numerical knowledge required to answer is not explicitly provided in the question.}
    \label{tab:mising_numerical_knowledge}
\end{table}

\begin{table}[ht]
    \centering
    
    \resizebox{\linewidth}{!}{
    % \hline
    \begin{tabular}{|l|l|r|}
    \hline
    % \toprule
    \textbf{Question} &\textbf{Knowledge Required} & \textbf{Answer} \\
    \hline
    \hline
    % \midrule
    \multicolumn{1}{|p{3.4cm}|}{Find the mass percentage of H in C6H6} & \multicolumn{1}{|p{2.1cm}|}{Mass of C is 12 units and mass of H is 1 unit} & 7.69\\
    % \hline
    % \midrule
    % Find molecular weight of NH4
    % &
    % molecular weight of N is 14 units and H 1 units 
    % &
    % 18
    % \\
    \hline
    \multicolumn{1}{|p{3.4cm}|}{How many units of H2 are required to react with 2 units of C2H4 to form 2 units of C2H6}
    & 
    \multicolumn{1}{|p{2.1cm}|}{H2 + C2H4 = C2H6}
    &
    2 \\
    % \midrule
    \hline
    
    % Find the number of units of H2 formed on combining 2 units of C2H6 and 4 units of O2
    % & 
    % C2H6 + 2O2 = 2CO2 + 3H2
    % &
    % 6 \\
    %\hline
    \multicolumn{1}{|p{3.4cm}|}{A car covers 912 meters in 19 seconds. If bike's speed is one fourth of the car. Find the distance covered by the bike in 4 seconds.}
    &
    \multicolumn{1}{|p{2.1cm}|}{distance travelled = speed * time}
    &
    48 \\
    \hline
    % \bottomrule
    \end{tabular}
    }
    \caption{Example questions where domain knowledge is required to answer a question.}
    \label{tab:chem}
\end{table}

\begin{table}[h]
    \centering
    
    \resizebox{\linewidth}{!}{
    \begin{tabular}{|l|l|}
    % \toprule
    \hline
        \textbf{QuaRel Question} &\textbf{Transformed Question} \\
    % \midrule
    \hline
    \hline
    % Tony has a beach ball that he loves to play with. He notices that the beach ball travels farther when he kicks it across the asphalt road than when he kicks it across the gravel road. If he kicks it with the same force in each situation, which road is more smooth? 
    % \newline
    % (A) gravel (B) \textbf{asphalt}

    % & Tony has a beach ball that he loves to play with. He notices that the beach ball travels for 6 meters when he kicks it across the asphalt road and 4 meters when he kicks it across the gravel road. If he kicks it with the same force in each situation, which road is more smooth? 
    % \newline
    % (A) gravel (B) \textbf{asphalt}
    % \\
    \hline
    \multicolumn{1}{|p{4cm}|}{A person wants to get shopping done quickly. They know that they can get through the checkout at big store faster than they can at small store. The store they go to to finish quickly is
    \newline
    (A) \textbf{big store} (B) small store}
    & \multicolumn{1}{|p{4cm}|}{ A person wants to get shopping done quickly. They know that they can get through the checkout at big store in 5 minutes whereas it can take 20 mintues at small store. The store they go to to finish quickly is
    \newline
    (A) \textbf{big store} (B) small store}
    \\
    \hline
    % \midrule
    
    \multicolumn{1}{|p{4cm}|}{Tina is racing her two dogs. Her greyhound is slim, her rottweiler is heavy. The dog that gets faster more quickly is the
    \newline
    (A) rottweiler (B) \textbf{greyhound}}
    &
    \multicolumn{1}{|p{4cm}|}{Tina is racing her two dogs. Her greyhound weighs 88 lbs and her rottweiler weighs 79 lbs. The dog that gets faster more quickly is the 
    \newline
    (A) \textbf{rottweiler} (B) greyhound}
    \\
    \hline
    % \midrule
    
    \multicolumn{1}{|p{4cm}|}{A golf ball has a smaller mass then a baseball. Which item has a weaker gravitational field?
    \newline
    (A) \textbf{golf ball} (B) baseball}
    &
    \multicolumn{1}{|p{4cm}|}{A golf ball has a mass of 78 grams and a baseball has a mass of 0.159 Kg. Which item has a weaker gravitational field?
    \newline
    (A) \textbf{golf ball} (B) baseball}\\
    
    \hline
    % \bottomrule
    \end{tabular}
    }
    \caption{Examples showing conversion of QuaRel questions to quantitative comparison questions}
    \label{tab:quarel}
\end{table}

\begin{table}[h]
    \centering
    
    \resizebox{\linewidth}{!}{
    \begin{tabular}{|l|l|}
    \hline
    % \toprule
    \textbf{Arithmetic Word Problem} &\textbf{Transformed Question} \\
    % \midrule
    \hline
    \hline

    \multicolumn{1}{|p{4cm}|}{Joan found 70 seashells on the beach. She gave Sam some of her seashells. She has 27 seashell left. How many seashells did she give to Sam ?
    \textbf{43}}
    & 
    \multicolumn{1}{|p{4cm}|}{Joan found 70 seashells on the beach . She gave Sam some of her seashells . She has 27 seashells left. She gave $\rule{1cm}{0.15mm}$ seashells to Sam.
    \textbf{43}}\\
    \hline
    \multicolumn{1}{|p{4cm}|}{Last week Tom had 74 dollars. He washed cars over the weekend and now has 86 dollars. How much money did he make washing cars ?
    \textbf{12}}
    & 
    \multicolumn{1}{|p{4cm}|}{Last week Tom had 74 dollars.  He washed cars over the weekend and made another 86 dollars. Tom has $\rule{1cm}{0.15mm}$ dollars now .
    \textbf{160}}\\
    \hline
    % \bottomrule
    \end{tabular}
    }
    \caption{Examples showing MAWPS questions and corresponding questions in Completion format }
    
    \label{tab:completion}
\end{table}

\end{document}